# Text-Based Approaches to Item Difficulty Modeling in Large-Scale Assessments: A Systematic Review

Sydney Peters, Nan Zhang, Hong Jiao,
Ming Li, Tianyi Zhou, Robert Lissitz

University of Maryland

## Abstract

Item difficulty plays a crucial role in test performance, interpretability of scores, and equity for all test-takers, especially in large-scale assessments. Traditional approaches to item difficulty modeling rely on field testing and classical test theory (CTT)-based item analysis or item response theory (IRT) calibration, which can be time-consuming and costly. To overcome these challenges, text-based approaches leveraging machine learning and language models, have emerged as promising alternatives. This paper reviews and synthesizes 37 articles on automated item difficulty prediction in large-scale assessment settings published through May 2025. For each study, we delineate the dataset, difficulty parameter, subject domain, item type, number of items, training and test data split, input, features, model, evaluation criteria, and model performance outcomes. Results showed that although classic machine learning models remain relevant due to their interpretability, state-of-the-art language models, using both small and large transformer-based architectures, can capture syntactic and semantic patterns without the need for manual feature engineering. Uniquely, model performance outcomes were summarized to serve as a benchmark for future research and overall, text-based methods have the potential to predict item difficulty with root mean square error (RMSE) as low as 0.165, Pearson correlation as high as 0.87, and accuracy as high as 0.806. The review concludes by discussing implications for practice and outlining future research directions for automated item difficulty modeling.

# Introduction

Large-scale assessments are often used for making high-stakes decisions such as identifying individuals that are qualified for grade promotion, college admission, or high-impact profession licensing or certifications such as doctors or nurses. These assessments must therefore adhere to professional standards for test development to ensure validity, reliability, and fairness (AERA, APA, & NCME, 2014). Accurately modeling item difficulty is one of the professional standards for item evaluation, because a balanced distribution of item difficulties is essential for developing assessments that will assure the same level measurement precision along the ability scale (AlKhuzaey et al., 2021; Parry, 2020). A test that contains too many easy or difficult items provides limited information at some other intervals along the ability scale, leading to less measurement precision at those ability levels (Hambleton, 1991).

## Item Difficulty Modeling

### Conventional Psychometric Approaches

The most common approach for estimating item difficulty has conventionally been conducted using item responses collected through field-testing. Newly created items are embedded in an operational test form for collecting item response data to estimate item parameters, but these items are not used for scoring (Benedetto, 2023). The student response data from these field-tested items are used to estimate the item difficulty using Classical Testing Theory (CTT) and Item Response Theory (IRT) frameworks (Hsu et al., 2018). CTT calculates the proportion of test-takers answering an item correctly, which is also known as an item's p-value. IRT uses a statistical model (either a logit or probit model) to represent the relationship among a test taker's latent ability, item parameters, and the likelihood that the test-taker answers the item correctly (DeMars, 2010). Despite its ability to yield accurate item difficulty estimates, this approach has been criticized for being time-consuming and costly (AlKhuzaey et al., 2024; Hsu et al., 2018). For instance, the process of field testing often takes several months to collect item response data. Second, IRT calibration typically requires administering items to several thousand examinees in large-scale assessments (Hambleton, 1991). Third, incorporating field-test items into operational test forms results in longer test durations, increasing the burden on both test-takers and administrators. Concerns are also raised about test-taking engagement, and item exposure, particularly in the development of high-stakes exams (Benedetto, 2023; Hsu et al., 2018; Loukina et al., 2016), as this could potentially compromise test security.

Literature also documented another approach for estimating item difficulty through expert ratings based on domain knowledge and experience, though this approach is seldom used in developing large-scale assessments. As expected, this method has been criticized for subjectivity. Conejo et al. (2014) found that content experts' ratings were often inconsistent and misaligned with psychometric estimates, with the correlation between human judgments and IRT-based item difficulty as low as 0.38. It is evident that experts may overestimate or underestimate test-takers' ability to answer items correctly, making their ratings unreliable as ground truth for item difficulty estimation, particularly in large-scale assessments. Additionally, this approach is impractical when dealing with a large volume of newly developed items, as it often requires multiple rounds of discussion to reach consensus, resulting in a time-consuming and costly process (Benedetto, 2023).

*Text-Based Approaches*

With the advances in natural language processing (NLP), there has been growing interest in automating the process of item difficulty prediction through the use of classical machine learning algorithms (e.g., Hsu et al., 2018; Loukina et al., 2016; Sano, 2015), deep learning models (Huang et al., 2017), transformer-based small language models (SLM; e.g., Benedetto, 2023; Tack et al., 2024) including BERT and its variants as well as large language models (LLM; e.g, Li et al., 2025; Rogoz & Ionescu, 2024) such as Phi and T5. This line of approach is also known as text-based difficulty modeling because only text of items and related meta-data such as item content label is utilized. Without collecting item response data, these methods have clear advantages when it comes to reducing time, cost, and subjective ratings.

Early literature was dominated by feature-based approaches that relied on manually crafted features informed by linguistic, cognitive, or psychometric theories related to item difficulty. Studies often combined multiple feature types, such as linguistic indicators and test metadata, to enhance item difficulty prediction. For example, Perkins et al. (1995) used a combination of linguistic (e.g., number of paragraphs per passage, propositional counts in both the passage and item stem) and test related (e.g., cognitive complexity of items) features. Some later articles utilized a large number of features to investigate if this improves item difficulty prediction. Loukina et al. (2016) developed a machine learning model to predict item difficulty in English language assessments by extracting over 100 hand-crafted linguistic features from the item stems, such as word length, syntactic complexity, and readability indices. The best result, using Random Forests, showed moderate correlation ($r = 0.50$) with ground truth EQ Delta, which was a standardized difficulty parameter under the framework of CTT. While these methods offer good interpretability, they rely heavily on manual feature extraction and therefore often fail to generalize well across different content domains.

Later studies began incorporating static word embeddings, numeric vector representations that capture semantic relationships among words. Examples include embeddings from count-based embedding techniques such as GloVe (Pennington et al., 2014) as well as predictive word embedding techniques like Word2Vec (Mikolov et al., 2013) and FastText (Bojanowski et al., 2017). The word embeddings served as automated feature extraction but lacked sensitivity to contextual word usage. For example, Hsu et al., (2018) applied word embedding techniques in predicting item difficulty into five levels for a large-scale social studies exam, the Basic Competence Test (BCTest). First, the data was preprocessed by splitting items into five parts: item stem, correct answer, and three distractors. Then, part of speech tagging was used to represent adverbs, conjunctions, prepositions, or auxiliary verbs. Next, Word2Vec was used to learn word embeddings from the text and convert the text into vectors. Word vectors were summed and semantic vectors were obtained for each item element. The cosine similarity was computed for each pair of item elements which were used in the prediction model (support vector machine) as semantic features. Results indicated that the proposed method achieved an overall accuracy of 34.74%, outperforming the baseline model's accuracy of 32.11%. This highlights the potential of word embedding techniques for estimating item difficulty, despite certain limitations in their performance. However, the extracted embeddings are static so they do not take into account the context and position of words, limiting their ability to capture contextual semantic relationships.

With the development of deep learning (LeCun et al., 2015), embeddings began to be extracted using deep neural networks. The generation of each vector considers how words and phrases interact within the context of the text, and therefore they are considered contextual embeddings. Based on model architecture, contextual embeddings can be produced by deep neural network models such as Bidirectional Long Short-Term Memory (BiLSTM) networks (e.g., Xue et al., 2020) or by models like ELMo (e.g., Ha et al., 2019; Xue et al., 2020), which utilize stacked BiLSTMs for contextual word representations. Another prominent category of contextual embeddings comes from transformer-based language models, such as BERT (Devlin et al., 2019), T5, and MPNet (e.g., Yousefpoori-Naeim et al., 2024). In general, transformer-based embeddings (e.g., from BERT) are more powerful, efficient, and flexible than embeddings generated by traditional models, such as BiLSTM.

Once features or embeddings are extracted, they were used as input data to train a classifier or a regressor, typically either a machine learning or a deep learning model (e.g., Convolutional Neural Network, CNN, He et al., 2021, Huang et al., 2017; Long Short-Term Memory, LSTM, Xue et al., 2020). For example, in predicting item difficulty for TOEFL reading comprehension items, He et al. (2021) employed a Convolutional Neural Network (CNN) as the prediction model. Their results showed that the proposed CNN approach achieved a root mean square error (RMSE) of 0.203, outperforming other methods such as regular CNN configuration (RMSE = 0.264) and a Support Vector Machine using TF-IDF representations (RMSE = 0.337).

Transformer-based language models (Devlin et al., 2019) can be broadly categorized into encoder-based smaller language models (SLMs) for text understanding and decoder-based large language models (LLMs) for text understanding and generation. The primary distinction between SLMs and LLMs also lies in the scale of model parameters: SLMs, such as BERT or DistilBERT, typically have hundreds of millions of parameters, whereas LLMs, like GPT-4o or Phi3, are trained on massive datasets and contain billions of parameters. As such, LLMs can perform a wide variety of tasks with less reliance on fine-tuning. Besides extracting embeddings, language models can be fine-tuned for item difficulty modeling. For instance, Tack et al. (2024) fine-tuned the RoBERTa-base model using only item text and expert-aligned skill labels. Their approach leveraged contextual embeddings to better capture semantic and syntactic patterns in the items, outperforming traditional feature-based models in accuracy.

LLM can help predict item difficulty in diverse ways. LLMs have been prompted to directly predict item difficulty or to generate input data or features that can be used in classical machine learning models or SLMs for item difficulty prediction. For example, Li et al. (2025) explored the use of GPT-4 to both predict item difficulty directly and to generate rationales for the difficulty of items. Further, LLMs with zero-shot have been used to generate answers to items (Rogoz & Ionescu, 2024), reasoning steps that lead to choosing the correct answer and distractors (Feng et al., 2025), or to model uncertainty (Zotos et al., 2025). These outputs were used as input data or features for other models to predict item difficulty. Additionally, Dueñas et al. (2024) leveraged LLMs to simulate diverse test-taker behaviors for item difficulty prediction, moving beyond content-related features. By extracting features from various prompting strategies including answering with justifications, yes/no answer reformulations, mutilated stems, and temperature-parameter variation, results showed that behaviorally derived features effectively capture item difficulty for a medical exam.

In summary, these examples illustrate how both feature-based, embedding-based, and language model-based automated methods can be applied to item difficulty prediction, with the latter offering enhanced performance through deeper contextual understanding. These developments suggest that modeling architecture for item difficulty prediction fundamentally shapes the type and depth of information leveraged in items. Considering the rapid advancement of NLP techniques, particularly the rise of transformer-based models including SLMs and LLMs, it is both timely and necessary to review recent modeling approaches that employ more advanced NLP and language models to synthesize the methods, datasets, and findings. Such a review can inform future developments in text-based item difficulty modeling by identifying untapped areas, limitations, and emerging trends.

**Previous Reviews on Text-based Item Difficulty Modeling**

The exploration of text-based approaches to model item difficulty has been ongoing for decades. A few review studies have synthesized and reflected on the research findings in a systematic way. Some of these studies concentrated on a specific domain and offered an overview of research in that area (e.g., Ferrara et al., 2022), while others focused on the models employed in the modeling process (e.g., AlKhuzaey et al., 2024; Benedetto et al., 2023). Ferrara et al. (2022) summarized 13 item difficulty modeling studies that focused on high-stakes reading comprehension exams. This review found that ordinary least squares regression were utilized in every study. Only two studies employed NLP techniques including Coh-metrix, Text Evaluator, or Psycho-Linguistic Measures of Assessment Content (PLIMAC). These findings highlight the emerging but still limited text-based methods for item difficulty prediction. Features found to be associated with item difficulty include the density of propositions, the level of concreteness or abstractness, the similarity between the language in the item stem and the passage, the vocabulary demands of the response options, and the plausibility of incorrect options.

More recent reviews included articles that employed advanced models, which rapidly emerged from the mid-2010s (e.g., AlKhuzaey et al., 2024; Benedetto et al., 2023). Benedetto et al. (2023) conducted a narrative review of the literature, synthesizing results from 38 studies published between 2015 and 2021. They provided a structured taxonomy to organize the methods and analyzed the most effective methods in different scenarios. For example, the educational domains were divided into language related and other content domains, due to the different sources of difficulty. While linguistic features play a key role in predicting item difficulty for language assessments, the topic or skill being assessed has a greater impact on item difficulty in other subject domains such as math and medicine. The results indicated that, for language assessment, simpler models (e.g., support vector machines and random forest) using linguistic features and readability measures can perform as well as or better than neural networks, including bidirectional long short-term memory (BiLSTM) and fully connected neural network (FCNN). However, in other subject domains, transformer-based models, generally achieve better performance. A reviewed study found BERT and DistilBERT outperformed random forest for a computer programming learning platform. The findings also highlighted a shift from classical machine learning models based on readability and word complexity to modern NLP-based approaches.

In contrast, AlKhuzaey et al. (2024) conducted a systematic review of 55 articles on item difficulty prediction without restricting the publication period, thereby covering studies from 1995 to 2022. Compared to previous reviews, they expanded the scope to provide an in-depth analysis of key aspects in terms of content domains, difficulty parameters, features, modeling approaches, input types, item formats, evaluation metrics, and publication trends over time. A key finding from the review is that linguistic features including syntactic and semantic features significantly influence item difficulty. The growing number of studies in psychometrics, educational psychology, and artificial intelligence highlights the increasing interest in automated item difficulty prediction.

Similarly, Luecht (2025), in a book chapter, reviewed studies published up to 2022 and presented item difficulty modeling as a central framework within assessment engineering. The chapter discussed the theoretical motivation and background for item difficulty modeling, rather than to provide a comprehensive review of models used in the literature. The author summarized item difficulty modeling into two paths: the strong theory pathway and the statistical control pathway. Early research mainly followed the strong theory pathway, which emphasizes designing items based on solid cognitive and learning theories about knowledge, skills, and experiences. With the development of machine learning and AI-based text analytics, research has shifted toward the statistical control pathway. This approach recognizes that it is not possible to fully control for each test-taker's unique background or every item-examinee interaction. Instead, researchers use tools like CohMetrix to select features that can help explain item difficulty, focusing on what works in practice rather than whether the features fit a specific cognitive theory. By examining these pathways, the review synthesized and organized years of item difficulty modeling research, tracing the shift from methods rooted in cognitive theory to newer AI and machine learning approaches that emphasize enhanced prediction performance.

Although AlKhuzaey et al. (2024) and Benedetto et al. (2023) synthesized the literature and provided detailed overviews of emerging automated item difficulty prediction methods, both reviews shared similar limitations. First, all reviewed articles were published no later than 2022. The book chapter by Luecht (2025) also only included a discussion of the evolution of item difficulty modeling through 2022. However, AI techniques such as language model-based approaches have advanced drastically in the past three years (2023-2025). Second, the reviews did not merely focus on large-scale assessments. Some reviewed studies were related to formative assessment for learning. Third, it is important to note that AlKhuzaey et al. (2024) included articles that used expert ratings as ground truth difficulty. However, this approach is not suitable for large-scale assessments because expert ratings can be subjective and often show poor agreement with empirical item difficulty estimates. Therefore, the present review focuses exclusively on studies where item difficulty, used as the gold standard for training models, was estimated from item response data.

In summary, prior reviews of text-based item difficulty modeling have provided in-depth theoretical context and analyzed key aspects such as trends in item types, input representations, modeling approaches, feature sets, evaluation criteria, and publication volume. However, even the most recent review (Luecht, 2025) covers research only up to 2022. With significant recent advances in NLP and language models, along with a surge in new studies, an updated synthesis is now needed.

**Purpose of the Study**

This review aims to highlight recent advances in the application of cutting-edge NLP technologies, particularly language model-based approaches for item difficulty prediction, with an emphasis on large-scale assessments. This review does not restrict content domain or year of publication. Another distinctive contribution of this paper is the reporting of model performance results, including the distribution of values across various evaluation metrics. Previous reviews (AlKhuzaey et al., 2024; Benedetto, 2023) focused on trends in the type of evaluation criteria but did not report model performance based on evaluation criteria because the results from studies using different item difficulty parameters, content domains, evaluation metrics, are not directly comparable. This is a valid concern; however, the synthesis of model performance can serve as reference points and benchmarks for future research.

The current review focuses on the use of machine learning and language models to predict item difficulty in large-scale assessment settings. The research questions guiding this review are presented as follows:
1. What text-based methods, especially advanced language model-based approaches, were applied to predict item difficulty?
2. What domains and item types were most frequently investigated?
3. Which text-based features (e.g., linguistic complexity, semantics, distractor similarity) were most frequently investigated in classical machine learning models?
4. Which evaluation criteria were used to assess model performance?
5. What was the distribution of model performance across the evaluation criteria?

## Methods

**Data Collection**

We conducted a comprehensive literature search for articles published through May 2025 across multiple databases, including Google, Google Scholar, IEEE Xplore, ArXiv, Scopus, Springer, and ERIC. Additional searches were performed on the websites of the National Council on Measurement in Education (NCME) and a relevant competition platform (i.e., the NBME Item Difficulty Prediction Competition) to locate papers submitted by participants. A Boolean search strategy was employed using keyword combinations in full text: *(item OR question) AND difficulty AND (AI prediction OR prediction using machine learning OR automatic prediction OR modeling)*.

After an initial screening based on titles, 93 articles were identified. Next, 17 articles were excluded based on the abstract and keywords, resulting in 76 articles. The full text of these articles was screened and 52 articles were excluded based on one or more of the following reasons: (1) the assessment was not large-scale, (2) the study focused on text complexity or readability rather than item difficulty, (3) the study did not focus on automated prediction based on item text, rather, they used ICC or response data to model the item difficulty, (4) the article was a review, or (5) the item difficulty parameter was not obtained from item responses from human test-takers, rather, the responses were from the AI agent. A total of 24 articles remained for in-depth analysis. Later, a forward hand-search was conducted to ensure that all related articles have been comprehensively included. For each included eligible article, we found all subsequent articles that cited it, and conducted another round of screening. In this procedure, 19 additional articles were found, and after screening following the same exclusion criteria listed above, 13 articles were included in the review. In total, 37 articles were coded and analyzed for this review, consisting of conference papers ($n = 20$), journal articles ($n = 7$), research reports ($n = 3$), pre-prints ($n = 5$), and master's or doctoral theses ($n = 2$). The full article selection process is summarized in Figure 1.

Since there could be more than one dataset analyzed in one paper, we treated the difficulty prediction on each dataset/assessment as a separate study. For instance, the difficulty modeling on RACE++ and ARC datasets in Benedetto (2023) were treated as two separate studies. The investigation of multiple item difficulty parameters within one paper were also considered as separate studies. For example, the modeling of the continuous IRT *b* parameter as well as the categorical difficulty level in Štěpánek et al. (2023) were also coded and analyzed respectively. Consequently, 46 studies resulted from the 37 articles. To differentiate the number of articles from the number of studies, we used *n* for the number of articles and *k* for the number of studies, hereafter. Count (*m*) is used for results when the count could exceed one for a single study (e.g., the exploration of multiple models within a study).

**Fig. 1** *Flowchart of Article Selection Process*

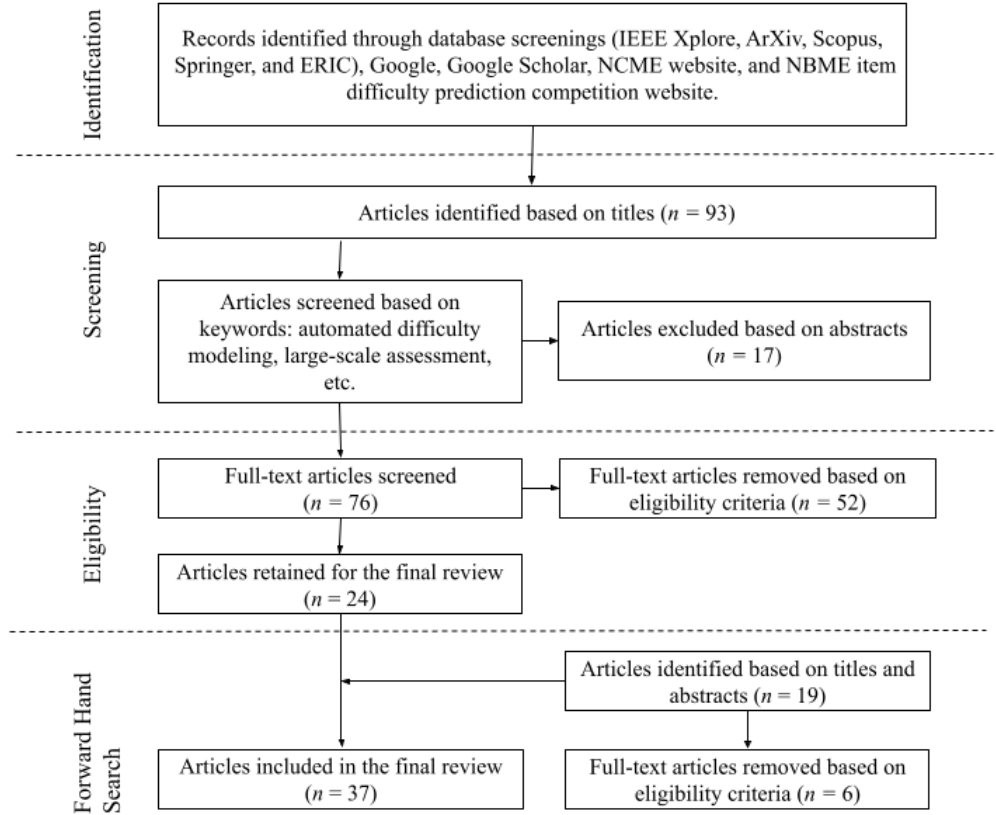

**Data Extraction**

We recorded the article information, dataset, and modeling methods for each selected article. Article Information included the article title, authors, and publication year; dataset contains assessment and dataset name, subject domain (e.g., language proficiency, math, science), item type (multiple-choice questions, short-answer questions, fill-in-the-blank, complete the forms/notes/table/flowchart/summary, complete the table, labeling a diagram/plan/map, classifying, true or false, and sorting), number of items; modeling approaches including item parameter predicted (e.g., p-value, IRT *b* parameter, EQ*Δ*), input data types, engineered features used in difficulty prediction, split proportion of training and test datasets, training models, model performance evaluation criteria (e.g., MSE, RMSE, *R* squared), and the evaluation outcomes. Two authors coded the articles independently, and any disagreements were adjudicated by a third party.

**Data Analyses**

A quantitative synthesis was conducted to examine method trends across the reviewed literature. Descriptive statistics were used to summarize key study characteristics, including publication year, item difficulty parameters, content domains, item formats, sample sizes, data splitting strategies, input configurations, feature types, modeling approaches, evaluation criteria, and model performance. Results were reported using count-based aggregation and percentages. When a single study contained multiple experiments using different item types, models, or input formats, each experiment was documented separately. Thus, a total of 160 model counts were recorded. Where relevant, trends in the literature were visualized using bar, line, and pie charts to enhance interpretability. Finally, model performance for each evaluation criterion with a sufficient amount of data across studies was summarized using descriptive statistics including minimum, maximum, median, mean, and standard deviation. No inferential statistical tests or meta-analytic modeling techniques were applied, given the heterogeneity of tasks, data sources, and outcome metrics across studies.

## Results

After coding and cleaning the data from the reviewed articles, we summarized the data of publication year, item difficulty parameter, content domains, item types, number of items, train and test dataset split, input data, features, models, evaluation criteria, and model performance. Thirty-seven articles were included in the final analysis, with 46 studies predicting the item difficulty. The following results are summarized based on the number of studies ($k$) or count ($m$) when multiple counts come from a single study.

**Data**
*Publication Year*

The distribution of articles over time reveals a marked increase in interest in automated item difficulty prediction for large-scale assessments beginning around 2013 (Figure 2). There were a few articles published in the late 1990s, followed by a long gap in the literature. It was picked up again in the early 2010s, with a dramatic increase in publications in recent years, peaking in 2024. This spike might be largely attributed to the Building Educational Applications (BEA) shared task on automated prediction of item difficulty and response time that was launched in June 2024 by National Board of Medical Examiners (NBME).

This pattern is consistent with the results from previous reviews which found research on item difficulty prediction has come in two waves (AlKhuzaey et al., 2024; Benedetto et al., 2023). The first wave, beginning in the mid-1990s, explored text-based methods including linear regression (Boldt & Freedle, 1996) and neural networks (Boldt, 1998; Bolt & Freedle, 1996; Perkins et al., 1995). This period was followed by a significant decline in publications from 1999 to 2012, with only a few relevant articles appearing during these years (Ayers & Junker, 2006; Dhillon, 2011; Fei et al., 2003; Hoshino & Nakagawa, 2010; Pérez et al., 2012; Wauters et al., 2012). However, these articles were excluded from the present review for one or more of the following reasons: they focused on task rather than item difficulty prediction, were not empirical studies, or did not pertain to large-scale assessments.

The second wave began in the early 2010s, marked by renewed interest in classic machine learning models (Beinborn et al., 2015; Loukina et al., 2016; Sano, 2015) and neural networks (Aryadoust, 2013; Aryadoust & Goh, 2014). AlKhuzaey et al. (2024) observed that the rise in studies may be directly linked to the surge of automated question generation research between 2014 and 2018 (Kurdi et al., 2021), as item difficulty modeling is crucial for assessing the quality of newly generated items. Another possible explanation is the increased adoption of computerized adaptive testing (CAT), in which items are dynamically selected according to the test-taker's estimated ability (AlKhuzaey et al., 2024). Several articles published in 2022 (Byrd & Srivastava, 2022; Xu et al., 2022) were identified but excluded, as they did not focus on large-scale assessments.

A steady rise in research on item difficulty modeling is evident from the growing number of publications since around 2013, compared to the mid-1990s and early 2000s (Figure 2). This trend reflects a broader surge of interest in applying artificial intelligence (AI) to educational measurement. To further explore this development, we analyzed the relationship between publication year and the models used. The results reveal a clear shift from classic machine learning methods to more advanced neural networks and transformer-based models. This transition highlights a growing preference among researchers for approaches capable of learning complex, non-linear representations directly from data, enabled by technological advances that facilitate the study of intricate relationships in language.

**Fig. 2** *Number of Articles Published Through Years 1995- (May) 2025*

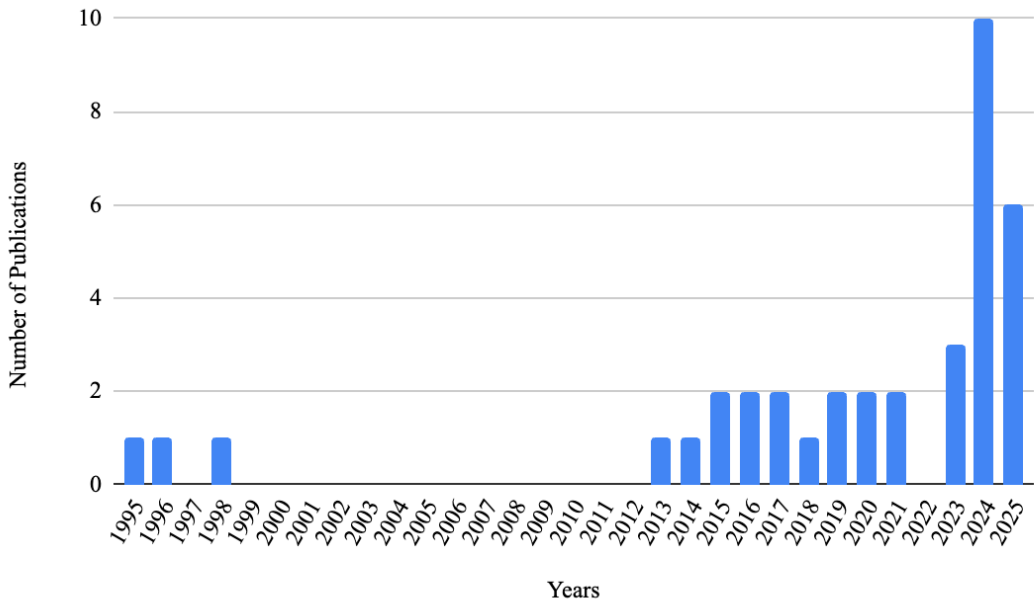

*Item Difficulty Parameters*

The type of item difficulty parameter used shapes the automated item difficulty prediction task. When item difficulty is represented as a continuous variable, prediction is framed as a regression problem, with performance evaluated by comparing predicted and empirical values (Benedetto, 2023). Conversely, if item difficulty is defined categorically (e.g., easy, medium, hard), the task becomes a classification problem (e.g., Hsu et al., 2018). In large-scale exams, most item difficulty studies predicted continuous values, aligning with the common practice of representing item difficulty using p-values or IRT *b* parameters.

Among the 46 reviewed studies, item difficulty was measured using different metrics, including IRT *b* parameter, p-values, or categorical levels. In addition to CTT p-value, defined as the proportion of examinees that get an item correct, some studies utilized the error rate as a difficulty parameter, q-value, defined as the proportion of examinees that get the item wrong. q-values range from 0 to 1, but a higher value indicates that the item is more difficult. In addition, the delta metric was used by ETS in the mid to late 1990s (Bolt 1998; Boldt & Freedle, 1996). Delta is calculated by transforming the proportion of incorrect responses into a normal deviate with a mean of 13 and a standard deviation of 4. Delta values are more likely to exhibit linear trends (Boldt, 1998). It is calculated by equation $\Delta = 13 - 4Z_p$, where $Z_p$ is the z-score (standard normal deviate) corresponding to an item's *p*-value.

As shown in Table 1, the most frequently reported item difficulty parameter was IRT *b* parameter ($k = 14$, 30.43%). Across studies, different IRT models were used including the Rasch model, two-parameter logistic (2PL) model, three-parameter logistic (3PL) model, graded response model (GRM), and generalized partial credit model (GPCM). The second most commonly used difficulty parameter was transformed p-value ($k = 11$, 23.91%). All of these studies came from the BEA shared task for predicting item difficulty for USMLE data, where the item difficulty parameter was the proportion of examinees that answered the item correctly with a linear transformation (Yaneva et al., 2024). The next most commonly used difficulty parameter was p-value ($k = 9$, 19.57%), followed by a categorical approach ($k = 5$, 10.87%), error rate ($k = 4$, 8.70%), and delta ($k = 3$, 6.52%).

Item difficulty parameters such as p-value, transformed p-value, error rate, and delta ($\Delta$) and EQ delta ($\Delta$) belong to the CTT category, adding up to 58.70% of the total reported item difficulty parameters. When grouped together, CTT-based parameters are used in 58.70% of studies, compared to 30.43% for the IRT b parameter. This suggests that the CTT framework is more commonly used in the reviewed studies. It is important to note that many of the reviewed articles ($k = 11$) came from the BEA shared task, which used transformed p-values. This dominance of CTT-based studies may be due to the popularity of shared-task competitions or the limited availability of public datasets with item difficulty values. Item difficulty levels were used 5 times, accounting for 10.87%.

**Table 1** *Counts and Percentages of Item Difficulty Parameters*

| Item Difficulty Parameter | Count (k) | Percentage (%) |
|---|---|---|
| IRT Model (*b*) | 14 | 30.43 |
| *p*-value | 9 | 19.57 |
| Transformed *p*-value | 11 | 23.91 |
| Error Rate | 4 | 8.70 |
| Delta | 3 | 6.52 |
| Categorical | 5 | 10.87 |
| Total | 46 | 100 |

However, CTT item difficulty measures have several shortcomings compared with IRT item difficulty measures. The former exhibited greater variation across different samples (e.g., female vs. male; high ability vs. low ability). The item difficulty parameter estimates based on the IRT theoretical framework were invariant across the different independent groups, and also the item difficulty parameter estimates for IRT were invariant across groups with varying sample sizes (Adedoyin et al., 2008). Overall, IRT offers the advantage of sample-invariant estimates, thus improving generalizability across examinee populations.

*Subject Domains*

Of the 46 reviewed studies, the test content domains included language proficiency ($k = 23$, 50.00%), medicine ($k = 15$, 32.61%), math ($k = 4$, 8.70%), science ($k = 2$, 4.35%), analytical reasoning ($k = 1$, 2.17%), and social studies ($k = 1$, 2.17%). Figure 3 depicts the distribution of content domains in these studies. The dominance of language proficiency and medicine is likely due to publicly available data. Several studies utilized the same USMLE dataset that was released for the BEA shared task for item difficulty. Additionally, the prevalence of language proficiency assessments may be partially attributable to the ease of extracting linguistic features which supports and aligns with the use of machine learning and language models.

**Fig. 3** *Distribution of Subject Domains*

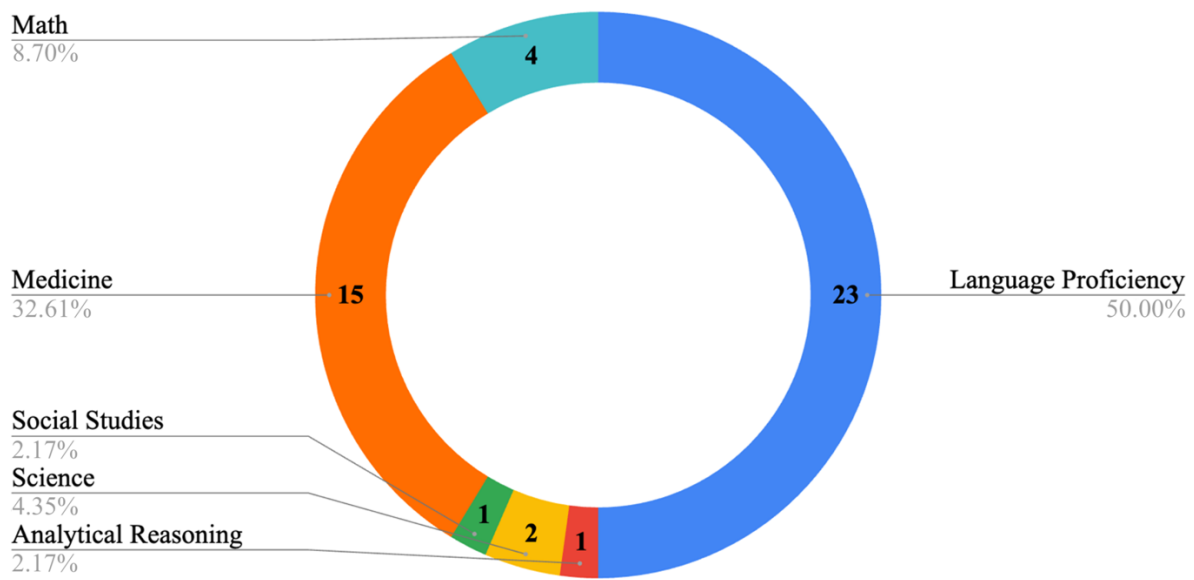

*Item Types*

A total of 60 item types were identified across the reviewed studies, as more than one item type could be examined within a single study. Multiple choice (MC) items accounted for the largest proportion, appearing 38 times (63.33%). Fill-in-blank items were reported eight times (13.33%), constructed-response items were reported four times (6.67%), and matching was reported twice (3.33%). Six item types (10.00%) were categorized as "Other." These included "complete the forms/notes/table/flowchart/summary", "complete the table", "labeling a diagram/plan/map", "classifying", "true or false," and "sorting", each of which had fewer than 2 occurrences. Finally, one article consisting of two studies did not specify the item types (3.33%). In general, the MC item types were dominant in item difficulty modeling.

**Fig. 4** *Distribution of Item Types*

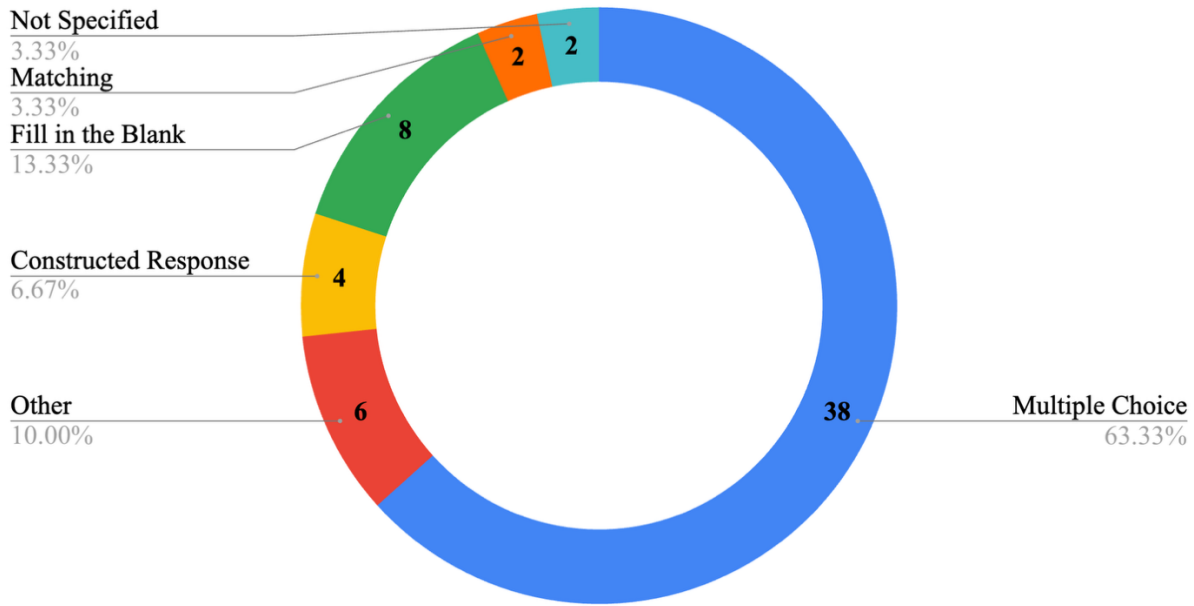

*Number of Items*

The number of items in each dataset was summarized into several ranges as shown in Table 2. The number of items varied widely in size, ranging from 100 to 106,210 items, demonstrating significant variability across studies. Most studies used items between 500 and 2,000, while only two studies utilized over 30,000 items with 106,210 items in RACE++ (Benedetto, 2023) and 30,817 items in IFLYTEK (Huang et al., 2017). It is important to note that counts are based on studies using unique datasets. Several articles utilized multiple datasets, each considered as a unique study. These include Beinborn et al. (2015) who used four datasets, and both Benedetto (2023) and Feng et al. (2025) who used two datasets. The K–5 mathematics and reading assessments in Razavi and Powers (2025) were treated as two separate datasets based on content domain. Štěpánek et al. (2023) reported two studies with the same dataset, which was counted only once, while Zotos et al. (2025) included three studies but utilized only two datasets. Full names of all abbreviated datasets are provided in Appendix A.

With regard to the number of items in each subject domain, the language proficiency domain included as few as 14 items (University of Duisburg-Essen German test) to as many as 106,210 items (RACE++) though both are outliers. This demonstrated the large variability in the number of items used for model development for language assessments. For studies in the medical domain, the number of items was consistently greater than 500. This is primarily due to the availability of the BEA shared task dataset presented in June 2024, which included 667 retired multiple-choice items from the USMLE (SIGEDU, 2024). Eleven studies used data from this competition (Bulut et al., 2024; Dueñas et al., 2024; Fulari & Rusert, 2024; Gombert et al., 2024; Li et al., 2025; Rodrgio et al., 2024; Rogoz & Ionescu, 2024; Tack et al., 2024; Veeramani et al., 2024; Yousefpoori-Naeim et al., 2024; Zotos et al., 2025). Before the competition, three studies utilized different USMLE item pools, each with a substantially larger number of items: 5,918 (Ha et al., 2019), 12,038 (Xue et al., 2020), and 18,000 items (Yaneva et al., 2020).

**Table 2** *Distribution of the Number of Items and Related Content Domain*

| Range | Count | Content Domain | Number of Items |
|---|---|---|---|
| 0-100 | 8 | Language | TU En (39), TU Fr (40), DE Ger (82), DE Ger (14), Czech Matura (40), IETLS (40), TOEFL (29), NAEP (21) |
| 100-500 | 7 | Language | MET (322), TOEFL (213), 3F Dutch (182) |
| | | Math | CHPCEE-M (263), APT (317), Eedi (327) |
| | | Science | KS2 (216) |
| 500-2,000 | 17 | Language | Brown NA Library (1,496), CMCQRD (793), NYSTP (1,076), ELA (750) |
| | | Medicine | USMLE - 11 Studies (667) |
| | | Social Studies | BCTEST (570) |
| | | Analytical Reasoning | GRE (1,457) |
| 2,000-5,000 | 4 | Language | Duolingo (3,826), K-5 Reading (2,606) |
| | | Medicine | ABP GP & GP-MOC (4,784) |

| | | Math | K-5 Math (2,564) |
| --- | --- | --- | --- |
| 5,000-10,000 | 3 | Language | ELP test (9,834) |
| | | Science | ARC (6,888) |
| | | Medicine | USMLE (5,918) |
| 10,000-30,000 | 3 | Language | TOEFL (12,416) |
| 30,000+ | 2 | Medicine | USMLE (12,038), USMLE (18,000) |
| | 2 | Language | IFLYTEK (30,817), RACE++ (106,210) |

**Model Training**
*Training and Test Dataset Split*

    To develop different models for item difficulty prediction, a dataset is often split into training, validation, and test datasets. The training set refers to the data that is used for models to learn patterns, relationships, and structures in the data, the validation dataset is used to fine-tune the model to achieve the best performance, whereas the test set is used to evaluate the trained model's performance on unseen data. Not all studies explicitly presented the percentage of data used for validation. Thus, training and validation datasets were combined for studies with three splits. The frequencies of the splitting proportions are presented in Figure 5. Several studies experimented with different train/test data split proportions (Benedetto, 2023; Bulut et al., 2024; Huang et al., 2017). As such, the total count of train/test data splits was 50. Across all studies, the best results were consistently achieved with the largest training datasets. Train/test data splits are not applicable to every model type and when it is applicable, it was not always explicitly reported. Thus, the total count from Figure 5 is not equal to the number of studies reviewed. All train and test splits are reported as percentages of the dataset.

    A wide variety of dataset splits were employed across studies, ranging from 40% for training and 60% testing, to 95% for training and 5% testing. The most common dataset split was 70% for training and 30% for testing ($m = 14$, 28.00%). For the most commonly studied dataset, the BEA shared task, the dataset consisted of 667 items, where 466 (70%) items were used for training, and the other 201 items (30%) were used for testing (Yaneva et al., 2024). The next most common data split was 80/20 ($m = 6$, 12.00%), followed by 50/50 ($m = 3$, 6.00%), 90/10 ($m = 3$, 6.00%), and 95/5 ($m = 3$, 6.00%). All other train-test splits were reported in only two studies or fewer, collectively accounting for 24.00%. Nine studies (18.00%) did not report the train and

test data split (Beinborn et al., 2015; Bolt & Freedle, 1996; He et al., 2021; Hsu et al., 2018; Sano, 2015; Xue, 2025), among which four were from Beinborn et al. (2015).

**Fig. 5** *Frequency Distribution of the Training and Testing Data Split Ratio*

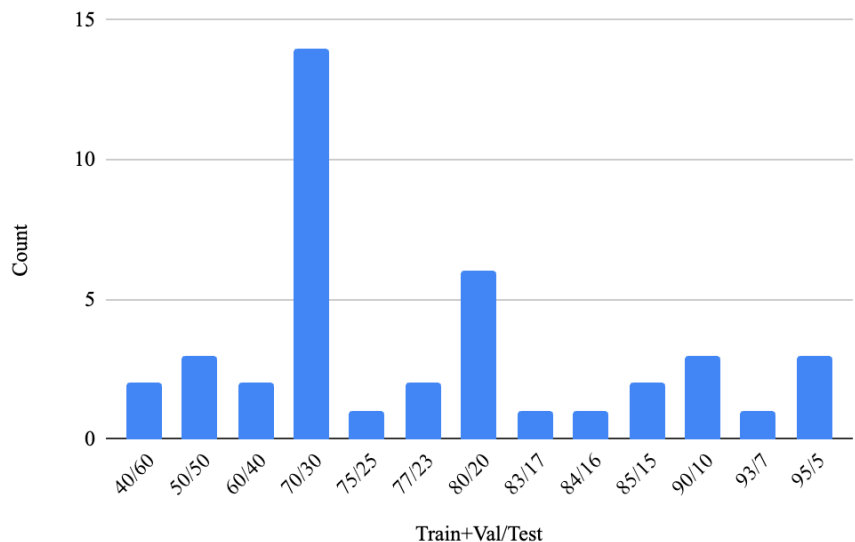

*Input Text Sources*

      The input data used to train the model refers to the original, unprocessed components of the item (e.g., item stem, correct answer, distractors, and stimulus like passage, tables, and figures when applicable). In addition, augmented data such as rationales generated by LLMs are also treated as input data. Table 3 summarizes the different combinations of item components that were used as input text to the item difficulty prediction models. Some studies experimented with multiple input text data, and each combination was counted once per study, yielding a total count of 74. The most common combination of item components used as input was Item Stem + Answer + Distractors ($m = 18$, 24.32%), followed by Item Stem Only ($m = 9$, 12.16%), then by Item Stem + Answer ($m = 6$, 8.11%). Some articles from the language proficiency tests included reading passages. Specifically, using Item Stem + Passage + Answer + Distractors was common ($m = 9$, 12.16%), as well as Item Stem + Passage ($m = 7$, 9.46%). Item Stem + Passage + Answer was reported once ($m = 1$, 1.35%).

      Four studies experimented with input combinations that did not utilize the item stem (Loukina et al., 2016; Rogoz & Ionescu, 2024; Tack et al., 2024; Xue et al., 2020). These inputs included answer text only, answer choice (letter of the option) only, text of all options including the correct answer, text of the distractors only, and passage only ($m = 6$, 8.11%). One early study (Boldt, 1998) did not utilize the raw item text as input, but rather input variables (coded item characteristics) derived from the item content (e.g., logical structure, object type) ($m = 1$, 1.35%).

      In addition, LLMs have been used to augment the input data ($m = 6$, 8.11%). Rogoz and Ionescu (2024) prompted three LLMs (i.e., Falcone-7B, Meditron-7B, and Mistral-7B) to answer each item by choosing one correct answer. The answer from the LLM was added to the input data. Veeramani et al. (2024) use LLMs (i.e., Mistral-7B, Llama-7B, and Gemma-7B) to

annotate named entity recognition (NER) and semantic role labeling (SRL). The outputs from these LLMs are then used as input data to predict item difficulty with the following components: item stem, answer, distractors, NER, and SRL. Other studies used GPT-4 to generate rationales for the difficulty of items (Li et al., 2025) or to generate reasoning steps to reach the correct answer and misconceptions or errors that lead to choosing each of the distractor options (Feng et al., 2025).

In sum, utilizing all item components appears to be the most frequently used input text source for item difficulty modeling. Richer input data lead to improved model performance. As demonstrated in Li et al., (2025), data augmentation in general can enhance model performance in item difficulty modeling. The authors experimented with two types of data augmentation: augmentation-on-the-fly, which generates augmented training samples dynamically during training to prevent overfitting, and augmentation with distribution balancing which generates augmented samples for underrepresented difficulty levels. The second method is proposed to enhance model performance by helping the model to better learn patterns from the full spectrum of difficulty levels. The best result from this study came from the BERT model with an ensemble of the two data augmentation methods, effectively leveraging the strengths of both.

**Table 3** *Counts and Percentages of Input Data*

| Input Data | Count | Percentage |
|---|---|---|
| Item Stem Only | 9 | 12.16 |
| Item Stem + Answer | 6 | 8.11 |
| Item Stem + Answer + Distractors | 18 | 24.32 |
| Item Stem + Passage | 7 | 9.46 |
| Item Stem + Passage + Answer | 1 | 1.35 |
| Item Stem + Passage + Answer + Distractors | 9 | 12.16 |
| Answer Text Only, Answer Choice, Text of Options, Distractors Only, Passage Only | 6 | 8.11 |
| Item stem + Figures or Item Stem + Oral Text + Tables/Charts | 2 | 2.07 |
| **LLM Augmented Data** Item Stem + LLM Answers + Answer, Item Stem + Answer + Distractors + NER + SRL, Item Stem + Answer + Distractors + LLM Rationales, Item Stem + Answer + Distractors + LLM Reasoning | 6 | 8.11 |
| Coded Item Characteristics | 1 | 1.35 |
| Not Specified | 9 | 12.16 |
| **Total** | 74 | 100 |

*Note.* Named Entity Recognition (NER) and Semantic Role Labeling (SRL) were generated from LLMs

*Features*

Features used for modeling can be generally classified into three categories, i.e., hand-crafted features, embeddings, and LLM generated features. Hand-crafted features are usually linguistic features. In some articles (e.g., Beinborn et al., 2015; Loukina et al., 2016), more than 50 linguistic features were extracted to predict item difficulty. Item or test metadata are sometimes used as features. For in-depth analysis, this review classified all features into five categories: linguistic features, item metadata, static embeddings, contextual embeddings, and LLM generated features. A total of 132 features were collected from 46 studies.

**Linguistic features** comprised 59.85% (79 feature counts) of the total number of features, including

(1) *Lexical features* including the number of words, the frequency of words, and word lengths, and Coh-Metrix features (e.g., Causal Connectives, Temporal Connectives, Anaphoric Pronoun Reference);
(2) *Syntactic features* including sentence count, average sentence length, part of speech, use of conjunctions, clauses, passive voice, number of noun or verb phrases;
(3) *Morphological features* including word stems, and lemmas; number of lemmas;
(4) *Semantic features* including the semantic similarity between item stem and choices such as cosine similarity, and semantic role labeling, which captures the meaningful roles of words or phrases in a sentence;
(5) *Readability indices* including features on type-token-ratio (vocabulary diversity), Flesch Reading Ease, Flesch-Kincaid Grade Level, ARI, Gunning FOG Index, Coleman-Liau Index, Linsear Write Formula, Dale-Chall Readability Score;
(6) *Content specific features* extracted from a content unique basis, such as the number of text-based numerical values for math.

**Item metadata** made up 21 (15.91%) feature counts, including

(1) *Cognitive complexity* including DOK levels of items;
(2) *Content standards* including curriculum standards, content tags;
(3) *Expert ratings* by human subject matter experts to assign the difficulty levels (usually in three levels or five levels);
(4) *Item characteristics* such as item types, the number of items per reading passage, the number of choices per item for MC items;

**Static embeddings** were used 8 times (6.06 %), including

(1) *Count-based embeddings* from GloVe;
(2) *Predictive embeddings* from Word2Vec.

**Contextual embeddings** were used 18 times (13.64%), including.

(1) *Deep learning model based embeddings* such as ELMo.
(2) *Word-level embeddings* such as BERT-base, DistilBERT, RoBERTa-base, ELECTRA, ALBERT, or MPNet.

(3) *Sentence-level SLM embeddings* from Sentence BERT, Longformer, Longformer, Clinical-longformer, S-PubMedBert-MS-MARCO.
(4) *LLM Embeddings* extracted from LLMs such as Llama.

**LLM generated features** was used 6 times (4.55%). GPT-3.5, GPT-4o, Llama, Falcon, Meditron, Mistral, Gemma, Qwen, Phi, and Yi have been used to generate features in the following categories:

(1) *Simulated responses:* item responses, length of justification, item performance when options converted to yes/no types, item responses when parts of the questions removed, item responses when temperature changed,
(2) *Cognitive features:* skill difficulty, distractor quality, and cognitive load
(3) *Response entropy:* how evenly the model distributes its predicted probabilities across all available choices; low entropy indicates high confidence in a single option, while high entropy reflects uncertainty and thus higher difficulty.
(4) *response latency:* the time the model takes to produce an answer, indicating processing difficulty
(5) *Proxy of Uncertainty* of LLMs:
   (a) First-token probabilities captures the model's initial confidence by measuring the likelihood it assigns to the first predicted token (e.g., a selected option)
   (b) Choice order sensitivity evaluates whether the model's answers change based on how the answer options are ordered, reflecting potential positional bias
   (c) Justification length refers to the number of tokens in the model's explanation, often used as a proxy for reasoning complexity or uncertainty.

These findings are consistent with the findings reported by AlKhuzaey et al. (2024) that 54% of all extracted features across studies were hand-crafted linguistic features, compared to only 8% derived from contextual embeddings. In comparison, the hand-crafted linguistic features were also dominant in our review (59.85%).

**Table 4** *Feature Types and Examples*

| Feature Types | Counts (%) | Subcategories | Examples |
| --- | --- | --- | --- |
| **Hand-crafted Features** | | | |
| Linguistic features | 79 (59.85%) | Lexical features | Word count, word length, complex word incidence, average sentence length Coh-metrix features (CNCCaus, CNCTemp, CRFANPa, CRFAO1, CRFCWO1, DESWLlt), TF-IDF embeddings, Bag of words |
| | | Syntactic features | Part of speech counts, sentence count, average sentence length, part of speech, use of conjunctions, clauses, passive voice, number of noun or verb phrases |
| | | Morphological features | Word stems, and lemmas; number of lemmas |
| | | Readability indices | Flesch Reading Ease, Flesch-Kincaid Grade Level, Gunning FOG Index |
| | | Semantic features | Cosine similarity semantic role labeling |
| | | Content specific features | Mathematical features: numerical values, number of text-based numerical values |
| Item Metadata | 21 (15.91%) | Cognitive complexity | DOK levels |
| | | Content standards | Curriculum standards, content tags |
| | | Expert rating | Human experts rating on difficulty levels |

|  | Item characteristics | Number of items per reading passage, item type |
|---|---|---|
| **Embeddings** | | |
| Static Embeddings 8 (6.06%) | Count based | Glove |
|  | Predictive | Word2Vec |
|  | Deep learning models | ELMo |
| Contextual Embeddings 18 (13.64%) | Small Language Model (word level) | BERT, BERT-clinical |
|  | Small Language Model (sentence level) | Longformer, Clinical-longformer |
|  | Large Language Model | Llama |
| **LLM Generated Features** | | |
| **LLM Generated** 6 (4.55%) | LLM Derived Features | Item responses simulated under different conditions, cognitive evaluation of items and options, response entropy, response latency, LLM uncertainty proxy |

**Figure 6** *Types of Features*

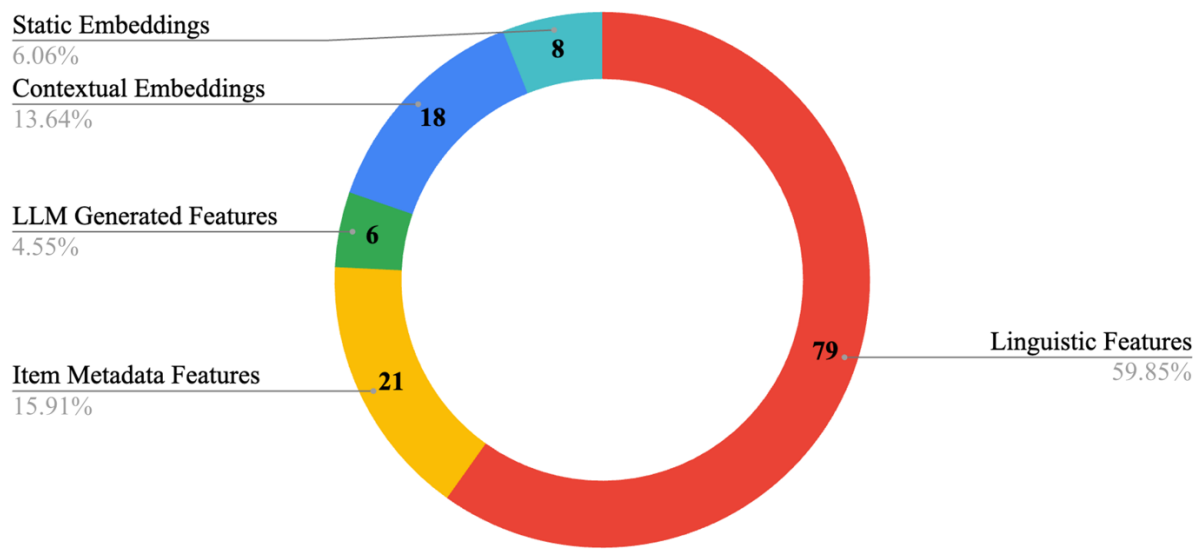

The hand-crafted features and item metadata features have been used for a long time with better interpretability. In contrast, the embeddings extracted from deep learning or language models is known for its black-box nature. The hand-crafted features do not necessarily perform worse than the contextual embeddings from BERT. For instance, Tack et al. (2024) used a large range of features such as item metadata including answer key, item type, exam step (i.e., the level of the exam), ordinal position of key within sequence of answer options, linguistic features (LIWC-22 features, lexical sophistication from TAALES), and BERT embeddings and found the hand-crafted features contributed more to the accuracy of difficulty prediction compared with the embeddings from BERT.

Recently, LLMs have been used as feature augmentation for item difficulty modeling. Recent study by Dueñas et al. (2024) prompted GPT-3.5 to simulate medical candidate responses to questions across diverse scenarios and extracted simulated response features from the model's behavior. Examples of extracted features include whether the model answered a question correctly, length of justification, performance when options were turned into yes/no questions, accuracy when parts of the question were removed to simulate skimming or missed details, and consistency under different temperature settings to simulate different test-taker behaviors (e.g., calm or anxious). Another study (Razavi & Powers, 2025) asked GPT-4o (2024-11-20 version) to reason item complexity and cognitive demand through targeted meta-cognitive questions. Cognitive features were extracted such as skill difficulty, distractor quality, and cognitive load, as well as subject domain specific features (e.g., use of visuals or multi-step reasoning for math and vocabulary and syntax complexity for reading). The features produced by GPT-4o in item difficulty prediction process was used to train random forest and gradient boosting models yielded high correlation of 0.87. Finally, Zotos et al. (2025) leveraged the uncertainty of LLMs from several families (i.e., Llama, Qwen, Phi, and Yi) and sizes (3B to 70B) to enhance item difficulty prediction. Uncertainty was measured in two ways: first token probabilities and choice order sensitivity. First token probabilities capture the model's initial confidence to answer the question by measuring the likelihood it assigns to the first token (e.g., a selected option) and choice order sensitivity evaluates whether the model's answers change based on how the answer options are ordered, reflecting potential position bias. This category of features is relatively new but they are promising to augment features utilizing LLMs' generative capacity.

To better understand the impact of feature types on model performance, prior studies compared different combinations of input features to assess their respective contributions to prediction accuracy. Rogoz and Ionescu (2024) and Tack et al. (2024) experimented with feature ablation and combination analyses to isolate the predictive power of different feature sets. Yi et al. (2024) applied a randomized feature order sensitivity test, which shuffled feature-label mappings in the training data to observe degradation in model performance. Though features are pivotal in modeling the difficulty parameters, more features did not necessarily lead to better model performance. Usually with the increased features, the dimension increases, which may lead to "curse of dimensionality" (e.g., Fulari & Rusert, 2024). Fulari and Rusert (2024) observed that linear regression performed markedly better when trained on a small set of engineered linguistic features (RMSE = 0.302) than on 768-dimensional contextual embeddings extracted from a fine-tuned BERT model (RMSE = 0.614). Some studies (Kapoor et al., 2025; Yi et al., 2024) applied Principal Component Analysis (PCA) to project high-dimensional features into more compact subspaces while preserving essential variance. This dimensionality reduction is particularly critical for models leveraging dense hand-crafted features (Yi et al., 2024) and embeddings (Kapoor et al., 2025). High-dimensional embeddings can introduce additional noise compared to a smaller set of carefully engineered features. Although embeddings are increasingly popular, linguistic features remain prevalent due to their interpretability.

*Models*

Among all reviewed studies, a total of 61 models have been explored 160 times, as it was common for studies to compare multiple models. In this section we only consider models that were used to directly predict item difficulty. The models were classified into three categories: classical machine learning models ($m$ = 94, 58.75%), deep learning models ($m$ = 20, 12.50%), and transformer-based language models, which includes both SLMs and LLMs ($m$ = 46, 28.75%). Among the transformer-based language models, 39 (84.78%) were SLMs and 7 (15.22%) were LLMs. Results are visually presented in Figure 7.

**Fig. 7** *Summary of Models*

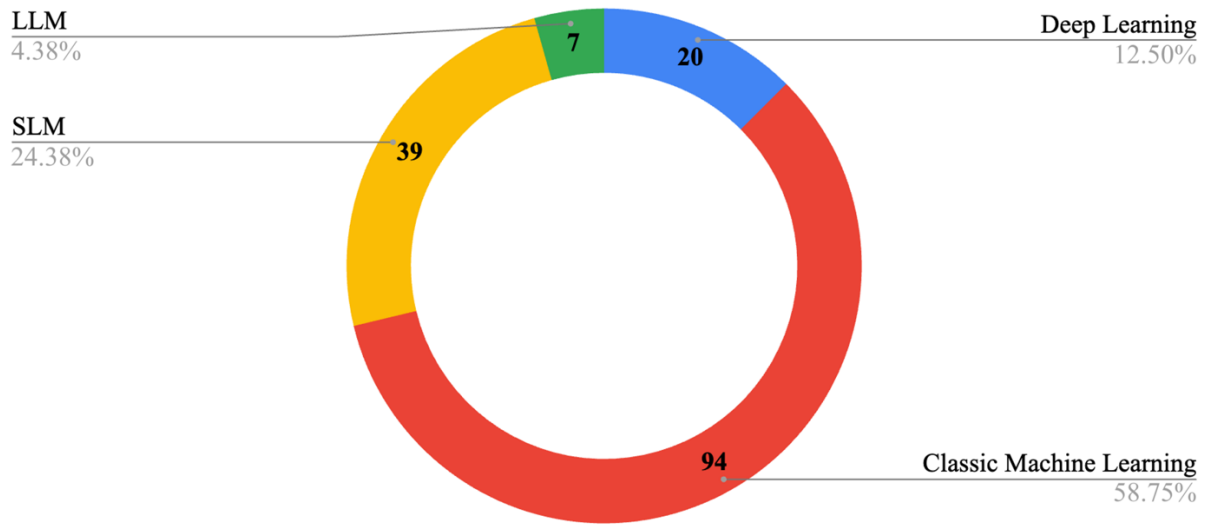

Classical machine learning models typically rely on engineered features. The classical machine learning models include the following types.

(1) *Linear and Penalized Regression Models* assume a linear relationship between the features and item difficulty. Regression models include Ordinary Least Square Regression, Partial Least Squares Regression, Elastic Net Regression, Lasso Regression,

Ridge Regression/Ridge (L2) Penalized Regression, and Linear Logistic Test Model (LLTM)[1].
(2) *Decision Tree-Based Models* use a tree-like structure of branching questions about features to predict item difficulty. Models include Classification and Regression Trees (CART), Classification Trees, Decision Tree Regression, Extra Trees, Random Forest, Regression Trees.
(3) *Probabilistic Models* utilize model probability distributions to model uncertainty to predict item difficulty. Models include Naive Bayes Classifier, Gaussian Processes, and Probabilistic Language Model.
(4) *Ensemble Learning Methods* combine the outputs of multiple models, often decision trees, to predict item difficulty. Models include AdaBoost, Cat-Boost, Gradient Boosting, Gradient Boosting Decision Trees, Light Gradient Boosting Machine, XGBoost, XGBoost-based SHAP Model.
(5) *Kernel and Distance-Based Models* measure similarity or distance between items based on their features to predict item difficulty. Models include k-Nearest Neighbors, Support Vector Machines.
(6) *Neural Network Based Models* use neural architectures but are typically contain one or no hidden layer. Models include Adaptive Neuro-Fuzzy Inference System (ANFIS), One Neuron Network (with no hidden layer), and Three-Layer Backpropagation Neural Network (with only one hidden layer).

Deep learning models use multiple hidden layers that mimic the functioning of human neurons to learn complex, non-linear representations from data but not based on attention mechanisms. These models use engineered features or embeddings. Although these models are better suited for capturing subtle patterns and contextual relationships in text, they come at the cost of interpretability and transparency. The explored deep learning models include
(1) *Basic Neural Network Architectures* that use interconnected layers that learn complex, nonlinear relationships between features and item difficulty*:* Artificial Neural Network (ANN), Multilayer-Perceptron (MLP), and Dense Neural Network.
(2) *Convolutional Neural Networks (CNNs) and Variants* that apply convolutional filters and attention mechanisms to improve item difficulty prediction*:* Convolutional Neural Network (CNN), Attention-based CNN (ACNN), Hierarchical Attention-Based CNN (HBCNN), Multi-Scale Attention CNN (MACNN), Temporal CNN (TCNN), Temporal Attention CNN (TACNN).
(3) *Bidirectional Long Short-Term Memory (Bi-LSTM)* that use recurrent neural network (RNN) architecture that processes input sequences in both forward and backward directions to capture contextual dependencies from both past and future inputs.

Transformer-based language models represent a specialized subset of deep learning in which the transformer architecture, characterized by self-attention mechanisms, is employed.

---

[1] LLTM is included in linear regression because it expresses item difficulty as a linear regression of features or embeddings within the IRT framework.

The self-attention mechanism contextualizes each word in the text by considering its relationship with all other words, regardless of position or distance. Transformer-based models consist of both SLMs and LLMs, where SLMs are defined as language models containing less than 1 billion parameters.

The reviewed SLM include:

(1) *BERT and its Variants*: BERT, BERT-ClinicalQA, Clinical-BERT, BioClinicalBERT, Bio_ClinicalBERT_emrqa, Bio_ClinicalBERT_FTMT, Clinical-BigBird, BioMedBERT, PubMedBERT, DistilBERT, ConvBERT, DeBERTa, RoBERTa, Electra, BioMedElectra

(2) *Long-Sequence Transformers*: specialized transformers that can process very long sequences of text, allowing difficulty prediction without the typical length limitations of other models: Longformer, Clinical-Longformer, Longformer-Base-4096, and BigBird.

(3) *GPT-2*: developed by OpenAI that generates text in an autoregressive manner by predicting the next token based on previous context. It ranges from 117 million to 1.5 billion parameters and demonstrated strong generalization across a variety of natural language tasks without task-specific training. The model being used is GPT-2 with 117 million parameters in the reviewed study.

(4) *T5 (Text-to-Text Transfer Transformer)*: developed by Google, T5 reformulates every NLP task as a text-to-text task. It uses an encoder-decoder architecture and is pre-trained on the C4 dataset. Widely used for summarization, translation, and classification.

The LLMs include:

(1) *GPT Family*: developed by OpenAI. Models reviewed include GPT-4, GPT-4o.

(2) *Llama Family*: developed by Meta. Models reviewed include Llama-7B

(3) *Mistral-7B*: a 7-billion parameter open-weight decoder-only model from the company Mistral AI (2023). Known for its strong performance relative to size, due to advanced training methods (e.g., grouped-query attention and sliding window attention).

(4) *Gemma-7B*: An open-weight model developed by Google DeepMind (2024). It is instruction-tuned and optimized for downstream tasks like text generation and reasoning, available for both research and commercial use.

(5) *Phi-3*: developed by Microsoft, Phi-3 is part of a family of small LLMs optimized for efficiency and performance on smaller hardware. It uses curriculum learning on filtered datasets for compact but strong generalization.

The classification of SLMs and LLMs are mainly size based. These models can be classified into encoder-only, decoder-only, or encoder–decoder models. Encoder-only models are designed to process the entire input sequence simultaneously and generate contextualized representations, good for classification or prediction. In contrast, decoder-only models operate in an autoregressive manner, ideal for text generation. Encoder–decoder models integrate both components for sequence-to-sequence tasks that require transforming one text into another. Most SLMs reviewed in this study are encoder-only models including BERT and its variants. GPT-2 is a decoder-only model with approximately 100 million parameters. T5 is an encoder–decoder model. The reviewed LLMs include decoder-only models such as GPT-4, GPT-4o, LLaMA, Mistral, Gemma, and Phi 3.

The BERT model was first explored for item difficulty modeling in large-scale assessments in 2021 (McCarthy et al., 2021). McCarthy et al. (2021) first demonstrated that BERT embeddings captured nuanced linguistic information beyond the static word embeddings by word2vec. Benedetto (2023) pioneered the end-to-end fine-tuning approach for the large-scale assessments by training both BERT and DistilBERT to predict MC item difficulty in reading and science. They found that transformer-based models consistently outperformed traditional linguistic, readability, embeddings from word2vec, and TF-IDF feature–based approaches even on relatively small datasets. However, in their study, even the smallest training set for RACE++ contained 4,000 items and the smallest training set for ARC contained 3,358 items, which is relatively large. DistilBERT matched BERT's performance at a cheaper cost.

Studies in 2024 began to explore more specialized transformer architectures and training strategies. For example, Zhao and Li (2024) fine-tuned RoBERTa and ELECTRA models on large-scale mathematics items, achieving a 5-7% reduction in RMSE over BERT baselines. Others (e.g., Gombert et al., 2024) experimented with multi-task learning, jointly predicting item difficulty and item content domain, and cognitive complexity.

*Models Studied Over Years.* To better understand how models evolved over the years, the number of articles that used each type of model in each publication year is plotted in Figure 8. Only models used for the direct prediction of difficulty were included and each model category was only counted once per article, regardless of different models within that category. For example, if an article employed multiple classical machine learning models including random forest and decision trees, only a single category, classical machine learning was recorded for this study.

The figure demonstrates several notable trends. Transformer-based language models did not appear in the item difficulty modeling literature until after 2020, aligned with the development of the BERT model and its variants. Statistical and classical machine learning models, by contrast, have been applied consistently over time, with a marked increase in use in recent years. Deep learning models have also been relatively consistent over time, but their prevalence appears to be on the decline since 2020. Collectively, these trends indicate a growing shift toward more advanced NLP and machine learning techniques in recent years, particularly with the increasing adoption of transformer-based architectures.

**Fig. 8** *The Trend of Studied Models over Years*

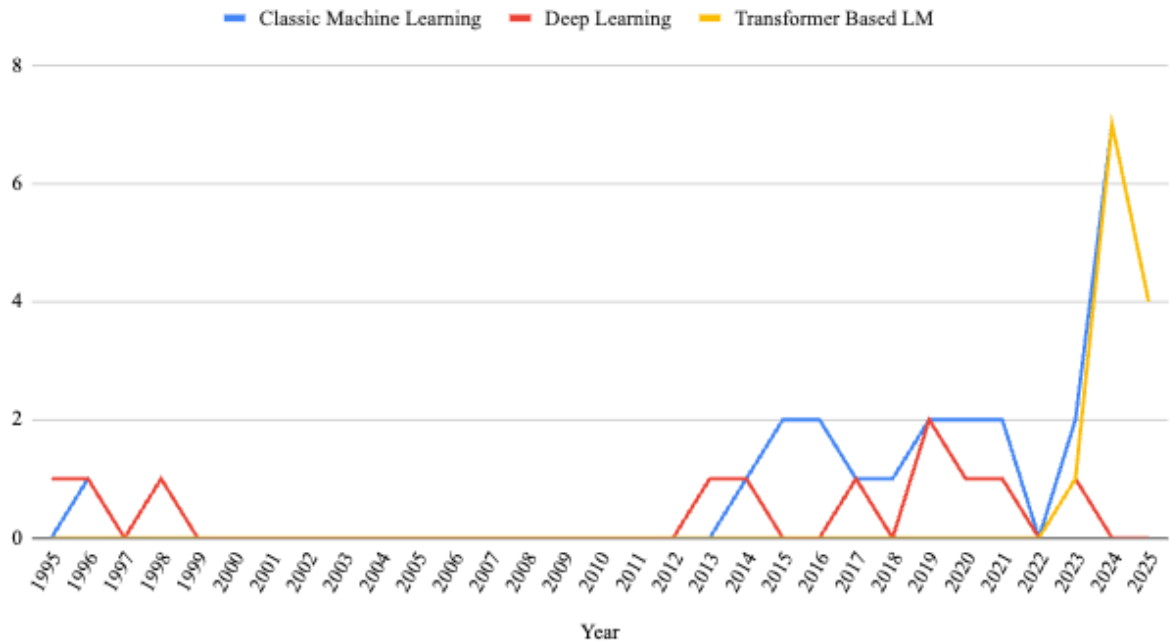

    Among the three categories of models, Transformer-based language models were virtually nonexistent before 2020 but have exhibited a significant upward trajectory since their introduction, making it feasible to encode rich semantic and syntactic cues that earlier architectures could not capture efficiently. Only one study used transformed models in 2021 and 2023, but their uses increased to eight studies in 2024 and five studies in 2025 (as of May).

    While researchers increasingly leverage state-of-the-art language models, classical machine learning techniques have retained momentum due to their transparency, interpretability, efficiency, and robustness with small sample sizes. Two factors likely explain this trend. First, many assessment datasets remain modest or even small in size, favoring classical machine learning models such as logistic regression, random forests, or gradient boosting that perform well with limited data. Second, with known features in such models, feature importance can be evaluated and facilitate the interpretability of item difficulty. This serves as an important validity evidence and informs item writing with targeted item difficulty levels.

    Finally, traditional deep learning models such as CNNs, Bi-LSTMs have been intermittently used beginning in 1995 and gaining moderate traction in 2019 to 2020. During this time, the use of deep learning models was approximately equal to the use of classical machine learning models. There has been a decline in the use of traditional deep learning models that coincided with the rise of transformer-based models. It is evident that models used for item difficulty modeling aligned with the advances in machine learning and NLP with consideration balancing sample size, interpretability, and computational cost and efficiency.

**Evaluation Criteria**

      Model performance was evaluated using different evaluation criteria, including measures of error, explained variance, correlation, and classification accuracy for item difficulty modeling. In total, 23 unique evaluation criteria with a total 96 counts were reviewed. Table 5 and Figure 9 summarize the overall distribution of evaluation criteria across all studies. Results showed that root mean square error (RMSE; $m = 28$, 29.17%), mean absolute error (MAE; $m = 8$, 8.33%), mean square error (MSE; $m = 5$, 5.21%), $R^2$ ($m = 13$, 13.54%), Pearson product moment correlation ($m = 17$, 17.71%), and Spearman rank order correlation ($m = 4$, 4.17%) are the most commonly used metrics to evaluate model performance for item difficulty prediction in the large scale assessment setting. Other metrics such as accuracy, F1 scores, recall, and precision had only been employed three or less times. This pattern likely stems from the fact that many studies ($m = 43$) framed item difficulty prediction as a regression task, whereas only a few studies ($m = 3$) framed it as a classification task, using categorical difficulty levels. As a result, regression-based evaluation metrics were more commonly employed than classification-based metrics. Although the datasets (RACE++ and ARC) used in Benedetto (2023) are labeled with categorical difficulty levels, these levels were first assigned discrete numeric values (0, 1, or 2 for RACE++ and 3 to 9 for ARC) used for regression analyses. The prediction models output continuous difficulty values which were converted to one of the categorical levels. Benedetto specifically stated that the item difficulty prediction was framed as a regression task and used some of the most common metrics in the literature (RMSE, $R^2$, and Spearman correlation) to evaluate model performance. Therefore, the studies from this article were considered regression tasks in the following summaries.

**Table 5** *Summary of Evaluation Criteria*

| Evaluation Criterion | Counts | Percentage (%) |
|---|---|---|
| **Error** | | |
|   RMSE | 28 | 29.17 |
|   MAE | 8 | 8.33 |
|   MSE | 5 | 5.21 |
|   Normalized MSE | 1 | 1.04 |
|   ME | 1 | 1.04 |
|   Bias | 1 | 1.04 |
|   Residual SD | 1 | 1.04 |
| **Explained Variance** | | |
|   R-Squared | 13 | 13.54 |
|   R-Squared Adjusted | 1 | 1.04 |
| **Correlation** | | |
|   Pearson | 17 | 17.71 |
|   Spearman | 4 | 4.17 |
|   Kendall | 1 | 1.04 |
| Test-Retest Reliability  | 1   | 1.04 |
| **Accuracy**             |     |      |
| Exact                    | 3   | 3.13 |
| Adjacent                 | 2   | 2.08 |
| F1                       | 1   | 1.04 |
| Recall                   | 1   | 1.04 |
| Precision                | 1   | 1.04 |
| Cross-Entropy            | 1   | 1.04 |
| **Others**               |     |      |
| Coefficient of Efficiency| 1   | 1.04 |
| Degree of Alignment      | 1   | 1.04 |
| Passing Ratio            | 1   | 1.04 |
| *Match                   | 2   | 2.08 |
| Total                    | 96  | 100  |

*Note*. RMSE = Root mean square error, MAE = Mean absolute error, MSE = Mean square error, ME = Max error of prediction, Residual SD = Residual Standard Deviation,

**Fig. 9** *Distribution of Evaluation Criteria*

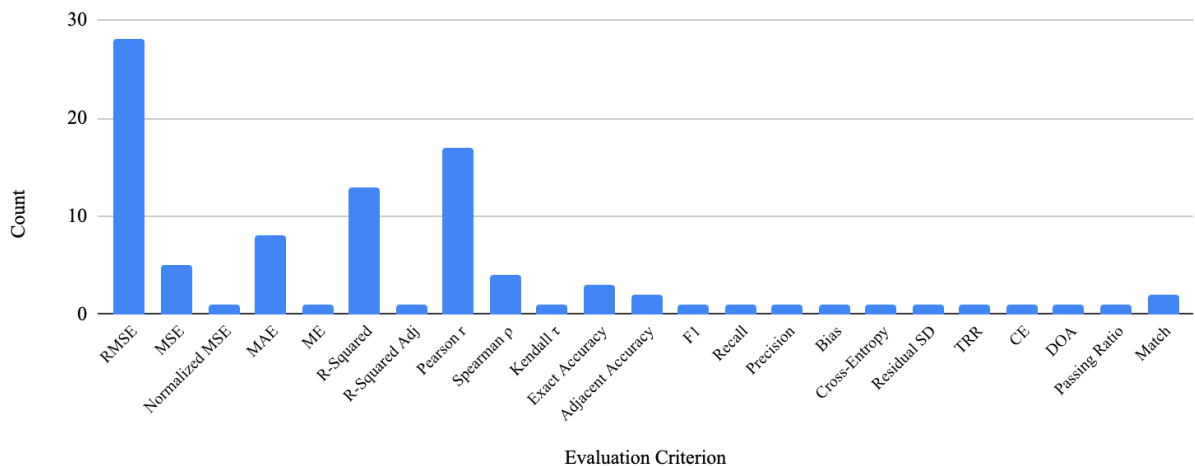

Note. TRR = Test-retest reliability, CE = Coefficient of efficiency, DOA = Degree of alignment

Table 6 summarizes the distribution of evaluation criteria for the 43 studies that framed item difficulty prediction as a regression task. The total count of evaluation criteria was 88, with 18 unique metrics. The most common evaluation metric used for prediction errors was RMSE ($m = 28$, 32.82%), followed by MAE ($m = 8$, 9.09%), and MSE ($m = 5$, 5.68%). Another common metric was $R^2$ to quantify the proportion of variance the model explains in item difficulty prediction ($m = 13$, 14.77%). This was usually coupled with a correlation metric that measures the strength of the association between the predicted and the empirical item difficulty. The most common type of correlation was Pearson product moment correlations ($m = 17$, 19.32%), followed by Spearman rank order correlations (*m* = 4, 4.55%), and Kendall's Tau, a non-parametric measure of the strength of association between two ranked variables (*m* = 1, 1.14%). Spearman and Kendall correlation are rank based metrics, they can be applicable for regression-based tasks to evaluate monotonic relationships between continuous variables. Such measures are often used when certain assumptions (e.g., linearity or normality) were not met for Pearson correlation. In addition, Feng et al. (2025) introduced a new evaluation criterion called "match" (*m* = 2, 2.27%). Match is a ranking metric that uses the predicted continuous difficulty values for each item to determine the relative ranking of each pair of multiple-choice items. The value of the metric is calculated as the percentage of item pairs that are correctly ranked. Finally, less frequently used evaluation criteria included normalized MSE, ME, bias, residual standard deviation, adjusted $R^2$, Kendall correlation, test-retest reliability, coefficient of efficiency, degree of alignment, passing ratio, and cross-entropy. Each of them was only used once (*m* = 1, 1.14%).

**Table 6** *Summary of Evaluation Criteria for Regression Tasks*

| Evaluation Criterion | Counts | Percentage (%) |
|---|---|---|
| **Error** | | |
|   RMSE | 28 | 31.82 |
|   MAE | 8 | 9.09 |
|   MSE | 5 | 5.68 |
|   Normalized MSE | 1 | 1.14 |
|   ME | 1 | 1.14 |
|   Bias | 1 | 1.14 |
|   Residual SD | 1 | 1.14 |
| **Explained Variance** | | |
|   R-Squared | 13 | 14.77 |
|   R-Squared Adjusted | 1 | 1.14 |
| **Correlation** | | |
|   Pearson | 17 | 19.32 |
|   Spearman | 4 | 4.55 |
|   Kendall | 1 | 1.14 |
|   Test-Retest Reliability | 1 | 1.14 |
| **Others** | | |
| Coefficient of Efficiency | 1 | 1.14 |
| Degree of Alignment | 1 | 1.14 |
| Passing Ratio | 1 | 1.14 |
| Cross-Entropy | 1 | 1.14 |
| *Match | 2 | 2.27 |
| Total | 88 | 100 |

Table 7 summarizes the distribution of evaluation criteria for the 3 studies that framed item difficulty prediction as a classification task. The total count of evaluation criteria was 8, with 5 unique metrics. Exact accuracy was used to evaluate model performance for every study in the classification setting ($m = 3$, 37.50%) and occasionally coupled with adjacent accuracy ($m = 2$, 25.00%), that quantifies accuracy in classifying items in the category immediately above or below the correct difficulty level. Less frequently used model evaluation criteria included F1 scores, recall, and precision. Each of them was only used once ($m = 1$, 12.50%).

**Table 7** *Summary of Evaluation Criteria for Classification Tasks*

| Evaluation Criterion | Counts | Percentage (%) |
|---|---|---|
| Accuracy | | |
|     Exact | 3 | 37.50 |
|     Adjacent | 2 | 25.00 |
| F1 | 1 | 12.50 |
| Recall | 1 | 12.50 |
| Precision | 1 | 12.50 |
| Total | 8 | 100 |

**Model Performance**

It was common for studies to compare different models using multiple evaluation criteria. Table 8 provides a summary of model performance in terms of the evaluation criteria. For example, if "Model 1" had a lower RMSE and lower Pearson correlation and "Model 2" had a higher RMSE and a higher Pearson correlation within a study, the RMSE from Model 1 and the Pearson correlation from Model 2 were used in the summary table. Evaluation criteria that were used two times were reported using the minimum and maximum values. While the table provides a summary of the distribution of values for the most commonly used evaluation criteria, it is important to highlight that these values cannot be directly compared across studies given these studies covered different subject domains and used different item difficulty metrics such as IRT $b$-parameters, CTT p-values, or classification of discrete difficulty levels. It is expected that the summarized results for various evaluation criteria provide a sense of what constitutes a "typical value".

For RMSE, the summary was made for $p$-value, transformed $p$-value, and the Rasch $b$ parameter respectively. The RMSE for studies using p-value ranged from 0.165 to 0.268 ($N = 6$, $M = 0.216$, $SD = 0.035$), while RMSE for studies using transformed p-values ranged from 0.253 to 0.308 ($N = 10$, $M = 0.291$, $SD = 0.018$). Again, all of the studies that used the transformed p-value difficulty parameter came from the BEA shared task for USMLE item difficulty prediction. One study that used the data from this competition did not report RMSE outcomes, instead they reported MSE. RMSE for studies using the Rasch model $b$ parameter ranged from 0.354 to 1.295 ($N = 8$, $M = 0.740$, $SD = 0.297$). Other IRT models were used only once. The RMSE for the difficulty parameters from other IRT models were as follows: 0.781 for 3PL/GPCM, 2.26 for delta, 0.372 for categorical with three levels, and 1.535 for categorical with seven levels. The Pearson correlation varied greatly across studies from 0.040 to 0.870 ($N = 17$, 0.545, $SD = 0.225$). R-squared values varied substantially across studies from 0.208 to 0.788 ($N = 13$, $M = 0.478$, $SD = 0.200$). The average R-squared value of 0.478 reflects moderate model performance while the highest reported value reached 0.788.

For classification, the range for exact accuracy also varied greatly across studies from 0.325 to 0.806 ($N = 3$, $M = 0.567$, $SD = 0.241$). However, the moderate to very high adjacent accuracy values ($N = 2$, 0.65 and 0.982) indicate that even when the model's prediction is not exactly accurate, oftentimes it is one category away from the expected difficulty level. Yi et al. (2024) was the only classification study in this review that used three difficulty levels - easy, medium, and hard. They reported outcomes for exact accuracy, F1 scores, recall, and precision. Both studies that utilized the adjacent accuracy metric predicted item difficulty into five levels (Hsu et al., 2018; Štěpánek et al., 2023).

**Table 8** *Summary of Model Performance*

| Evaluation Criterion | Count | Min | Max | Median | Mean | SD |
|---|---|---|---|---|---|---|
| **Regression Tasks** | | | | | | |
| RMSE | | | | | | |
|    Based on p-value | 6 | .165 | .268 | .214 | .216 | .035 |
|    Based on transformed p-value | 10 | .253 | .308 | .297 | .291 | .018 |
|    Based on Rasch model | 8 | .354 | 1.295 | .693 | .740 | .297 |
| MSE | 5 | .013 | .521 | .064 | .203 | .227 |
| MAE | 7 | .185 | .58 | .24 | .307 | .159 |
| Correlation | | | | | | |
|    Pearson | 17 | .04 | .87 | .55 | .545 | .225 |
|    Spearman | 4 | .25 | .790 | .496 | .508 | .221 |
| R-Squared | 13 | .208 | .788 | .525 | .478 | .200 |
| Match | 2 | .757 | .780 | - | - | - |
| **Classification Tasks** | | | | | | |
| Accuracy | | | | | | |
|    Exact | 3 | .325 | .806 | .569 | .567 | .241 |
|    Adjacent | 2 | .65 | .982 | - | - | - |

Among the evaluation criteria, 14 criteria were only used once, and their empirical values are listed in Table 9. The table also includes RMSE that were reported only once across the literature for item difficulty parameters for the 3PL/GPCM, delta, and categorical with either three or seven levels. These values summarize model performance in terms of less commonly item difficulty metrics and evaluation criteria.

**Table 9** *Summary of Model Performance for Uncommon Evaluation Criteria*

| Evaluation Criterion | Outcome |
|---|---|
| **Regression** | |
| RMSE | |
|     Based on 3PL/GPCM | 0.781 |
|     Based on Delta | 2.26 |
|     Based on categorical (3 levels) | 0.372 |
|     Based on categorical (7 levels) | 1.535 |
| Normalized MSE | 0.542 |
| ME | 7.180 |
| Bias | 0.001 |
| Residual SD | 0.16 |
| R-Squared Adjusted | 0.383 |
| Kendall Correlation | 0.423 |
| Test-Retest Reliability | 0.59 |
| Coefficient of Efficiency | 0.981 |
| Degree of Alignment | 0.665 |
| Passing Ratio | 0.42 |
| Cross Entropy | 0.53 |
| **Classification** | |
| F1-Score | 0.942 |
| Recall | 0.941 |
| Precision | 0.944 |

*Note.* The ME outcome is based on the Delta difficulty parameter. Decimal places are reported exactly as presented in the respective studies.

## Summary and Discussion

The purpose of this review was to synthesize text-based item difficulty prediction research in large-scale assessments using machine learning and deep learning models. A total of 46 studies from 37 articles were included for this review. To address the research questions, summaries were synthesized about the dataset (e.g., subject domains, item types), modeling methods, features, evaluation criteria, and model performance. The major findings are presented below, highlighting modeling approaches and domain-specific results. The contribution of this review, remained issues, and future directions are discussed.

### Major Findings

This review found two distinct waves of research interest in item difficulty modeling after the early exploration in the mid-1990s, a resurgence beginning in the early 2010s followed by a pronounced spike in 2024 largely driven by the BEA shared task using USMLE dataset. Reviewed studies explored different metrics of item difficulty, either CTT p-values or IRT-based *b*-parameters, with p-values more frequently used. Language proficiency and medical assessments were the most frequently studied subject domains. MC items were the most common item format, and the number of items used for model training varied substantially, from fewer than 100 items to over 100,000 items, highlighting the large variability in the number of items used with obvious outliers at two ends.

The more frequently used input data for model training include item stem, correct answer, distractors, and more recently LLM-generated input data. Hand-crafted features included linguistic features and item meta-data features with better interpretability. Embedding features include static embeddings and contextualized embeddings. Despite the increasing adoption of advanced language models, hand-crafted features remained competitive in predictive performance and offer the advantage of greater interpretability.

Given the lag of applications of the latest language model, classical machine learning models were most prevalent and exhibited consistent use over time, though the adoption of transformer-based models has accelerated in recent years. As item difficulty modeling can be either a regression or classification task, RMSE, $R^2$, and correlation are common metrics for regression analyses, while exact and adjacent accuracy are commonly used for item difficulty level classification. Although model performance across studies are not directly comparable due to differences in datasets, item difficulty scale, and evaluation criteria, the best-performing models suggested the promise of text-based approaches to item difficulty modeling, reaching accuracy as high as 0.806 and RMSE as low as 0.165.

Further, despite the growing popularity of transformer-based models, model selection should take sample size and quality of the data into consideration. Even though LLMs are increasingly used in recent studies, the SLMs may outperform LLMs for some datasets. However, fine-tuning large models on small datasets is prone to overfitting (Rogoz & Ionescu, 2024). For example, training an SVR on BERT embeddings showed better generalization than fine-tuning BERT a model due to small sample sizes (Rogoz & Ionescu, 2024).As suggested by Li et al. (2025) and Rogoz and Ionescu (2024), a dataset of 500 items is considered a small dataset. This review in general supported a training sample size of 1000 items for maintaining adequate technical quality.

This review found that the subject domain plays a role in the importance of features in item difficulty prediction. Linguistic features are of great importance in predicting item difficulty for language proficiency assessments, and remain relevant for other subject domains such as medicine, math, and science. In addition, content-related features are also of importance for other subject domains.

The findings from this review present several contributions to educational assessment. First, this review synthesized the models and the factors that needed to be taken into account in developing text-based approaches for item difficulty prediction. This review presents information about the input data, sample size, models, evaluation criteria, and model performance. In general, richer input from item stem, correct answer, and distractors and a large training dataset increased model performance.

*Issues Remained*

The reviewed studies are limited in data access and variation of data from different assessment programs. A trade-off between data accessibility and data privacy is always challenging. On one hand, increasing public access to high-quality educational assessment items can accelerate research progress by enabling replication, benchmarking, and innovation for item difficulty modeling. On the other hand, making these datasets widely available raises valid concerns about data security. Public release of assessment items and student responses may compromise test security. Furthermore, as LLMs are often trained on web-scraped or publicly available text, given the limited access to educational assessment data, there is a high risk that LLMs trained on the limited assessment data may introduce potential bias in the model performance without representation from assessments for different subject domains. Though training local SLMs and LLMs is a potential solution, sample size turned out to be a potential issue for text-based item difficulty modeling.

CTT-based item difficulty parameters, p-values or its transformed versions were widely used in the reviewed studies. Item p-values are group dependent measures (Adedoyin et al., 2008; Hambleton & Jones, 1993). Given the wide use of IRT b parameters over CTT *p*-values in large-scale assessments, it is expected more studies will explore the prediction of the IRT *b* parameter. For comparison purposes, it is highly recommended that common benchmarking metrics are used across studies to enable meaningful comparisons across studies. For example, evaluation criteria including RMSE, $R^2$, and correlation coefficients can be used across studies for regression analysis. The lack of standardization and consistent evaluation metrics hinders the synthesis of findings across studies. The issue of standardized benchmarks and evaluation criteria warrants attention. In this review, comparisons across studies are complicated by different evaluation criteria (prediction/classification errors, variance explained, or correlation) used by different studies. Establishing standardized benchmarking protocols and common evaluation metrics would facilitate valid comparisons across modeling approaches among studies, greatly benefiting the research community (Deng et al., 2020). Initiatives such as community-driven shared tasks and competitions could drive methodological standardization and foster a collaborative research ecosystem.

Despite the critical role of data preprocessing in model development, it remains under-reported in the reviewed studies. Preprocessing such as text normalization, tokenization, removal of noise data, and consistent handling of missing or extraneous information are essential for efficient feature extraction and model training. However, few studies provided detailed information of data preprocessing. Improved transparency in documenting data preprocessing is vital for feature-based approaches to support replication and cross-study synthesis.

Imbalance in item difficulty distributions is frequently observed in large-scale assessments. However, only one study addressed this issue (Li et al., 2025). The ignorance of imbalance issues in item difficulty distribution can substantially distort model performance, as predictive models may become biased toward majority difficulty levels while neglecting underrepresented ones. Li et al. (2025) explicitly documented imbalance of item difficulty, where the majority of items fell in the easy difficulty range, leaving difficult items underrepresented in the training data. To address this, their study proposed a distribution balancing augmentation strategy that generated more synthetic training items in sparsely populated difficult bins, ultimately yielding lower RMSE values. Their finding highlights the importance of examining and dealing with imbalance in item difficulty modeling.

***Limitations & Future Directions***

One limitation of this review is the potential bias introduced by the overrepresentation of papers from the BEA (Building Educational Applications) shared task and associated workshop. While these papers provide valuable insights and often reflect the state-of-the-art in item difficulty prediction, their prevalence in the dataset may have skewed the overall findings and model trends observed in this review. For instance, all of these studies used the same dataset with 667 items, and they typically used the same train/test dataset split. As such, the findings from this review may not fully generalize to broader or more diverse educational assessment contexts. Future reviews could benefit from a more balanced inclusion of studies using different large-scale assessment items from different subject domains to mitigate this bias.

Further, the predominance of MC items among the reviewed studies (63.33%) highlights a limitation in item types addressed by current item difficulty modeling research. This overrepresentation likely reflects the ubiquity of MC items in large-scale assessments. However, this dominance limits the generalizability of findings to other item types such as CR, fill-in-blank. Moreover, item types that require more complex interactions, such as labeling, matching, or completing structured tables, were rarely included (13.33%). Future studies should aim to capture diverse item types for difficulty modeling.

Text-based approaches to item difficulty prediction are expected to improve with the advances in NLP. Techniques developed for question- answering, summarization, and large reasoning models may offer promising adaptations for improving item difficulty prediction methods. Exploring transformer hybrid models or ensemble approaches that integrate predictions from multiple modeling paradigms (classical ML, deep learning, and transformers) could further increase accuracy and reduce errors. Such methods await more intensive exploration for solving issues and improving practices in large-scale assessment programs to save human and financial resources for field testing and item reviews.

Moreover, it is essential to consider model transparency and explainability from an educational measurement standpoint. The ability for test developers to interpret and act upon item difficulty predictions critically depends on the transparency of the underlying predictive models. Classical machine learning models often offer clear interpretability through explicit assumptions and straightforward feature importance metrics (Hastie et al., 2009). In contrast, deep learning and transformer-based models, although powerful, generally function as "black boxes," making their inner workings and decisions less transparent to test developers (Doshi-Velez & Kim, 2017). Researchers in the area of deep learning have tried to investigate the reasoning procedures from LLMs and use those for difficulty prediction and alignment (Zhao et al., 2024). Future research should prioritize approaches that enhance model interpretability, such as feature importance analysis and explainable artificial intelligence methods, to ensure that predictions remain actionable and meaningful in practical settings.

It is also worth noting the relatively late adoption of state-of-the-art models such as BERT in item difficulty prediction. Although BERT was introduced in 2018, its first use for item difficulty modeling was documented for 2021. This lag highlights the slow uptake of innovations from NLP. As AI technology rapidly evolve, proactively monitoring the developments in AI, continued and timely integration of cutting-edge AI technology are essential for advancing item difficulty modeling.

*Conclusion*

This review synthesized the methods used for text-based item difficulty modeling in large-scale assessments. It provided empirical evidence of the impact of design factors on model performance in predicting item difficulty based on item texts. This integration of educational measurement and machine learning turns out to be a promising approach to item difficulty modeling as part of assessment engineering, emphasizing computational efficiency and methodological innovation. Addressing these considerations will not only advance the field methodologically but also enhance practical utility in large-scale educational assessments.

Traditional methods of item difficulty estimation using an IRT model depend on item response data collection in either field testing or operational tests, which require extensive human labor, time consuming, and considerable financial investment (Hambleton & Jones, 1993; Lord & Novick, 1968). The automated methods reviewed in this report demonstrate high potential to alleviate these challenges. These methods can rapidly process and analyze large numbers of newly developed items before field-testing, significantly reducing the time from item writing to item reviews and evaluation (Bodily et al., 2022). For large-scale state assessments, this translates into streamlined item development processes, with reduced costs but enhanced quality engineered with the latest AI technology (Smith & Jalali, 2022).

# References

*References marked with an asterisk (*) indicate studies included in the meta-analysis.*

Adedoyin, O. O., Nenty, H. J., & Chilisa, B. (2008). Investigating the invariance of item difficulty parameter estimates based on CTT and IRT. *Educational Research and Reviews*, *3*(3), 83.

AlKhuzaey, S., Grasso, F., Payne, T.R., Tamma, V. (2021). A Systematic Review of Data-Driven Approaches to Item Difficulty Prediction. In: Roll, I., McNamara, D., Sosnovsky, S., Luckin, R., Dimitrova, V. (eds) Artificial Intelligence in Education. AIED 2021. Lecture Notes in Computer Science(), voll 12748. Springer, Cham.

AlKhuzaey, S., Grasso, F., Payne, T. R., & Tamma, V. (2024). Text-based question difficulty prediction: A systematic review of automatic approaches. International Journal of Artificial Intelligence in Education, 34(3), 862-914.

American Educational Research Association, American Psychological Association, & National Council on Measurement in Education. (2014). Standards for educational and psychological testing. American Educational Research Association.

*Aryadoust, V. (2013, April). Predicting item difficulty in a language test with an Adaptive Neuro Fuzzy Inference System. In 2013 ieee workshop on hybrid intelligent models and applications (hima) (pp. 43-50). IEEE.

*Aryadoust, V., & Goh, C. C. (2014). Predicting listening item difficulty with language complexity measures: A comparative data mining study. *CaMLA Work. Pap*, *2*, 1-16.

Ayers, E., & Junker, B. W. (2006, July). Do skills combine additively to predict task difficulty in eighth grade mathematics. Educational Data Mining: Papers from the AAAI Workshop. Menlo Park, CA: AAAI Press.

*Beinborn, L., Zesch, T., & Gurevych, I. (2015). Candidate evaluation strategies for improved difficulty prediction of language tests. In Proceedings of the Tenth Workshop on Innovative Use of NLP for Building Educational Applications (pp. 1–11). Denver, CO: Association for Computational Linguistics.

*Benedetto, L. (2023, June). A quantitative study of NLP approaches to question difficulty estimation. In International Conference on Artificial Intelligence in Education (pp. 428-434). Cham: Springer Nature Switzerland.

Bloom, B. S. (1956). *Taxonomy of educational objectives: The classification of educational goals*. Longmans.

Bodily, S. E., Kwon, M., & Hammond, T. (2022). Automated item analysis: Accelerating the pathway from item development to implementation. *Assessment in Education, 29*(4), 467–485.

Bojanowski, P., Grave, E., Joulin, A., & Mikolov, T. (2017). Enriching word vectors with subword information. Transactions of the association for computational linguistics, 5, 135-146.

*Boldt, R. F., & Freedle, R. (1996). Using a neural net to predict item difficulty. *ETS Research Report Series*, *1996*(2), i-19.

*Boldt, R. F. (1998). GRE analytical reasoning item statistics prediction study. *ETS Research Report Series*, *1998*(2), i-23.

*Bulut, O., Gorgun, G., & Tan, B. (2024). Item Difficulty and Response Time Prediction with Large Language Models: An Empirical Analysis of USMLE Items.

Byrd, M., & Srivastava, S. (2022, May). Predicting difficulty and discrimination of natural language questions. In Proceedings of the 60th Annual Meeting of the Association for Computational Linguistics (Volume 2: Short Papers) (pp. 119-130).

Clark, P., Cowhey, I., Etzioni, O., Khot, T., Sabharwal, A., Schoenick, C., & Tafjord, O. (2018). Think you have solved question answering? try arc, the ai2 reasoning challenge. arXiv preprint arXiv:1803.05457.

Conejo, R., Guzmán, E., Perez-De-La-Cruz, J. L., & Barros, B. (2014). An empirical study on the quantitative notion of task difficulty. Expert Systems with Applications, 41(2), 594-606.

Corbett-Davies, S., & Goel, S. (2018). The measure and mismeasure of fairness: A critical review of machine learning fairness metrics. *Journal of Machine Learning Research, 81*, 1–48.

Devlin, J., Chang, M. W., Lee, K., & Toutanova, K. (2019, June). Bert: Pre-training of deep bidirectional transformers for language understanding. In Proceedings of the 2019 conference of the North American chapter of the association for computational linguistics: human language technologies, volume 1 (long and short papers) (pp. 4171-4186).

DeMars, C. (2010). Item response theory. Oxford University Press.

Deng, J., Zhang, X., & Wang, L. (2020). A bottleneck in benchmarking: Public datasets for item difficulty modeling. *Educational Data Mining, 12*(1), 1–19.

Dhillon, D. (2011). Predictive models of question difficulty–A critical review of the literature. The Assessment and Qualifications Alliance, 21.

Doshi-Velez, F., & Kim, B. (2017). Towards a rigorous science of interpretable machine learning. *arXiv preprint arXiv:1702.08608*.

*Dueñas, G., Jimenez, S., & Ferro, G. M. (2024, June). Upn-icc at bea 2024 shared task: Leveraging llms for multiple-choice questions difficulty prediction. In Proceedings of the 19th Workshop on Innovative Use of NLP for Building Educational Applications (BEA 2024) (pp. 542-550).

*El Masri, Y. H., Ferrara, S., Foltz, P. W., & Baird, J. A. (2016). Predicting item difficulty of science national curriculum tests: the case of key stage 2 assessments. The Curriculum Journal, 28(1), 59–82.

Fei, T., Heng, W. J., Toh, K. C., & Qi, T. (2003, December). Question classification for e-learning by artificial neural network. In Fourth international conference on information, communications and signal processing, 2003 and the fourth pacific rim conference on multimedia. Proceedings of the 2003 joint (Vol. 3, pp. 1757-1761). IEEE.

*Feng, W., Tran, P., Sireci, S., & Lan, A. (2025). Reasoning and Sampling-Augmented MCQ Difficulty Prediction via LLMs. arXiv preprint arXiv:2503.08551.

Ferrara, S., Steedle, J. T., & Frantz, R. S. (2022). Response demands of reading comprehension test items: A review of item difficulty modeling studies. Applied Measurement in Education, 35(3), 237-253.

*Fulari, R., & Rusert, J. (2024, June). Utilizing Machine Learning to Predict Question Difficulty and Response Time for Enhanced Test Construction. In Proceedings of the 19th Workshop on Innovative Use of NLP for Building Educational Applications (BEA 2024) (pp. 528-533).

*Gombert, S., Menzel, L., Di Mitri, D., & Drachsler, H. (2024, June). Predicting item difficulty and item response time with scalar-mixed transformer encoder models and rational network regression heads. In Proceedings of the 19th Workshop on Innovative Use of NLP for Building Educational Applications (BEA 2024) (pp. 483-492).

*Groot, N. (2023). *Using Task Features to Predict Item Difficulty and Item Discrimination in 3F Dutch Reading Comprehension Exams* (Master's thesis, University of Twente).

Hambleton, R. K., Swaminathan, H., & Rogers, H. J. (1991). Fundamentals of item response theory (Vol. 2). Sage.

Hambleton, R. K., & Jones, R. W. (1993). Comparison of classical test theory and item response theory and their applications to test development. *Educational measurement: issues and practice*, *12*(3), 38-47.

Hastie, T., Tibshirani, R., & Friedman, J. (2009). *The elements of statistical learning: Data mining, inference, and prediction* (2nd ed.). Springer.

*Ha, L. A., Yaneva, V., Baldwin, P., & Mee, J. (2019). Predicting the difficulty of multiple choice questions in a high-stakes medical exam. In Proceedings of the Fourteenth Workshop on Innovative Use of NLP for Building Educational Applications (pp. 11–20). Florence, Italy: Association for Computational Linguistics.

*He, J., Peng, L., Sun, B., Yu, L., & Zhang, Y. (2021). Automatically predict question difficulty for reading comprehension exercises. In 2021 IEEE 33rd International Conference on Tools with Artificial Intelligence (ICTAI) (pp. 1398-1402).

*Hsu, F. Y., Lee, H. M., Chang, T. H., & Sung, Y. T. (2018). Automated estimation of item difficulty for multiple-choice tests: An application of word embedding techniques. Information Processing & Management, 54(6), 969–984.

Hoshino, A., & Nakagawa, H. (2010). Predicting the difficulty of multiple-choice close questions for computer-adaptive testing. Natural Language Processing and its Applications, 46, 279-292.

*Huang, Z., Liu, Q., Chen, E., Zhao, H., Gao, M., Wei, S., ... & Hu, G. (2017, February). Question Difficulty Prediction for READING Problems in Standard Tests. In *Proceedings of the AAAI conference on artificial intelligence* (Vol. 31, No. 1).

*Kapoor, R., Truong, S. T., Haber, N., Ruiz-Primo, M. A., & Domingue, B. W. (2025). Prediction of Item Difficulty for Reading Comprehension Items by Creation of Annotated Item Repository. arXiv preprint arXiv:2502.20663.

Kurdi, G., Leo, J., Matentzoglu, N., Parsia, B., Sattler, U., Forge, S., ... Dowling, W. (2021). A comparative study of methods for a priori prediction of MCQ difficulty. Semantic Web, 12(3), 449–465

LeCun, Y., Bengio, Y., & Hinton, G. (2015). Deep learning. nature, 521(7553), 436-444.

Levy, R., Smith, J., & Nguyen, P. (2021). Impact of train-test splits on model robustness in educational assessment models. *International Journal of Educational Technology in Higher Education, 18*(1), 45–62.

*Li, M., Jiao, H., Zhou, T., Zhang, N., Peters, S., Lissitz, R. (2025). Item Difficulty Modeling Using Fine-Tuned Small and Large Language Models. *Educational and Psychological Measurement*. (accepted).

Lord, F. M., & Novick, M. R. (1968). *Statistical theories of mental test scores*. Addison-Wesley.

Luecht, R. M. (2025). Assessment engineering in test design: Methods and applications. Taylor & Francis.

*Loukina, A., Yoon, S. Y., Sakano, J., Wei, Y., & Sheehan, K. (2016, December). Textual complexity as a predictor of difficulty of listening items in language proficiency tests. In *Proceedings of COLING 2016, the 26th International Conference on Computational Linguistics: Technical Papers* (pp. 3245-3253).

*McCarthy, A. D., Yancey, K. P., LaFlair, G. T., Egbert, J., Liao, M., & Settles, B. (2021). Jump-Starting Item Parameters for Adaptive Language Tests. Proceedings of the 2021 Conference on Empirical Methods in Natural Language Processing. Pp. 883-899.

Mehrabi, N., Morstatter, F., Saxena, N., Lerman, K., & Galstyan, A. (2021). A survey on bias and fairness in machine learning. *ACM Computing Surveys, 54*(6), 1–35.

Parry, J. R. (2020). Ensuring Fairness in Difficulty and Content among Parallel Assessments Generated from a Test-Item Database. Online Submission.

Pérez, E. V., Santos, L. M. R., Pérez, M. J. V., de Castro Fernández, J. P., & Martín, R. G. (2012, October). Automatic classification of question difficulty level: Teachers' estimation vs. students' perception. In 2012 frontiers in education conference proceedings (pp. 1-5). IEEE.

*Perkins, K., Gupta, L., & Tammana, R. (1995). Predicting item difficulty in a reading comprehension test with an artificial neural network. Language Testing, 12(1), 34-53.

*Qunbar, S. A. (2019). Automated item difficulty modeling with test item representations (ERIC No. ED601723). [Doctoral dissertation, The University of North Carolina at Greensboro].

*Razavi, P., & Powers, S. J. (2025). Estimating Item Difficulty Using Large Language Models and Tree-Based Machine Learning Algorithms. arXiv preprint arXiv:2504.08804.

*Rodrigo, A., Moreno-Álvarez, S., & Peñas, A. (2024, June). Uned team at bea 2024 shared task: Testing different input formats for predicting item difficulty and response time in

medical exams. In Proceedings of the 19th Workshop on Innovative Use of NLP for Building Educational Applications (BEA 2024) (pp. 567-570).

*Rogoz, A. C., & Ionescu, R. T. (2024). UnibucLLM: Harnessing LLMs for Automated Prediction of Item Difficulty and Response Time for Multiple-Choice Questions. *arXiv preprint arXiv:2404.13343*.

*Sano, M. (2015). Automated capturing of psycho-linguistic features in reading assessment text. Paper presented at the annual meeting of the National Council on Measurement in Education (NCME), Chicago, IL.

SIGEDU. (2024). BEA 2024 Shared Task: Automated Prediction of Item Difficulty and Item Response Time. https://sig-edu.org/sharedtask/2024

*Štěpánek, L., Dlouhá, J., & Martinková, P. (2023). Item difficulty prediction using item text features: Comparison of predictive performance across machine-learning algorithms. Mathematics, 11(19), 1-30.

*Tack, A., Buseyne, S., Chen, C., D'hondt, R., De Vrindt, M., Gharahighehi, A., Metwaly, S., Nakano, F. K., & Noreillie, A.-S. (2024). ITEC at BEA 2024 shared task: Predicting difficulty and response time of medical exam questions with statistical, machine learning, and language models. In Proceedings of the 19th Workshop on Innovative Use of NLP for Building Educational Applications (BEA 2024) (pp. 512–521).

*Trace, J., Brown, J. D., Janssen, G., & Kozhevnikova, L. (2017). Determining cloze item difficulty from item and passage characteristics across different learner backgrounds. Language Testing, 34(2), 151–174.

*Veeramani, H., Thapa, S., Shankar, N. B., & Alwan, A. (2024, June). Large Language Model-based Pipeline for Item Difficulty and Response Time Estimation for Educational Assessments. In Proceedings of the 19th Workshop on Innovative Use of NLP for Building Educational Applications (BEA 2024) (pp. 561-566).

Wauters, K., Desmet, P., & Van Den Noortgate, W. (2012). Item difficulty estimation: An auspicious collaboration between data and judgment. Computers & Education, 58(4), 1183-1193.

Xu, J., Wei, T., & Lv, P. (2022). SQL-DP: A Novel Difficulty Prediction Framework for SQL Programming Problems. International Educational Data Mining Society.

*Xue, K., Yaneva, V., Runyon, C., & Baldwin, P. (2020). Predicting the difficulty and response time of multiple choice questions using transfer learning. In Proceedings of the Fifteenth Workshop on Innovative Use of NLP for Building Educational Applications.

*Xue, M., Han, S., Boykin, A., & Rijmen, F. (2025, April). *Leveraging Large Language Models in Predicting Item Difficulty*. Paper presented at the annual meeting of the National Council on Measurement in Education, Denver, CO.

*Yaneva, V., Baldwin, P., & Mee, J. (2019, August). Predicting the difficulty of multiple choice questions in a high-stakes medical exam. In *Proceedings of the fourteenth workshop on innovative use of NLP for building educational applications* (pp. 11-20).

*Yaneva, V., Ha, L. A., Baldwin, P., & Mee, J. (2020). Predicting item survival for multiple-choice questions in a high-stakes medical exam. *In Proceedings of the Twelfth Language Resources and Evaluation Conference* (pp. 6812–6818). European Language Resources Association.

Yaneva, V., North, K., Baldwin, P., Ha, L. A., Rezayi, S., Zhou, Y., ... & Clauser, B. (2024, June). Findings from the first shared task on automated prediction of difficulty and response time for multiple-choice questions. In Proceedings of the 19th Workshop on Innovative Use of NLP for Building Educational Applications (BEA 2024) (pp. 470-482).

*Yi, X., Sun, J., & Wu, X. (2024). Novel feature-based difficulty prediction method for mathematics items using XGBoost-based SHAP model. Mathematics, 12(10), 1455.

*Yousefpoori-Naeim, M., Zargari, S., & Hatami, Z. (2024, June). Using machine learning to predict item difficulty and response time in medical tests. In *Proceedings of the 19th Workshop on Innovative Use of NLP for Building Educational Applications (BEA 2024)* (pp. 551-560).

Zhao, H., Yang, F., Lakkaraju, H., & Du, M. (2024). Opening the black box of large language models: Two views on holistic interpretability. *arXiv e-prints*, arXiv-2402.

*Zotos, L., van Rijn, H., & Nissim, M. (2024). Are You Doubtful? Oh, It Might Be Difficult Then! Exploring the Use of Model Uncertainty for Question Difficulty Estimation. arXiv preprint arXiv:2412.11831.

**Appendix A**

*Full Titles of Assessments and Datasets Referenced in Table 2*

| Abbreviation | Full Title or Dataset Elaboration |
|---|---|
| 3F Dutch | Reading Comprehension Exams for Dutch Vocational Education |
| ABP GP & GP-MOC | American Board of Pediatrics Certifying Examination in General Pediatrics |
| APT | Adult Proficiency Math Test |
| ARC | AI2 Reasoning Challenge Dataset (Clark et al., 2018) |
| BCTEST | Basic Competence Test |
| CHPCEE-M | Chen's Henan Provincial College Entrance Examination in Mathematics |
| CMCQRD | Cambridge Multiple-Choice Questions Reading Dataset |
| Czech Matura | Czech Matura Examination (English as a Foreign Language) |
| DE Ger | University of Duisburg-Essen German Prefix Deletion Test |
| ELA | English Language Arts Test |
| ELP Test | International English Language Proficiency Test |
| GRE | Graduate Record Examination |
| IELTS | International English Language Testing System |
| IFLYTEK | A dataset from iFlytek (KeDaXunFei), a Chinese technology company. Dataset collected from real-world standard tests for reading problems. |
| KS2 | Key Stage 2 Science Test |
| MET | Michigan English Test |
| NAEP | National Assessment of Educational Progress |
| NYSTP | New York State Testing Program Reading Comprehension Test |
| RACE++ | ReAding Comprehension dataset from Examinations |
| TOEFL | Test of English as a Foreign Language |
| TU En | Technische Universität Darmstadt English C-test |
| TU Fr | Technische Universität Darmstadt French C-test |
| USMLE | United States Medical Licensure Examination |